%% file: main_arxivV2.tex
\documentclass[10pt,twocolumn,letterpaper]{article}

\usepackage{cvpr}              


\usepackage{nicematrix}
\usepackage{graphicx}
\usepackage[accsupp]{axessibility}
\usepackage{xcolor}
\usepackage{makecell}
\definecolor{darkgreen}{rgb}{0.0, 0.5, 0.0}
\definecolor{darkblue}{rgb}{0.0, 0.4, 1}
\usepackage{amssymb}
\usepackage{pifont}
\usepackage{placeins}
\usepackage{balance}
\usepackage{flushend}
\newcommand{\cmark}{\ding{51}}%
\newcommand{\xmark}{\ding{55}}%

\usepackage{algorithm}
\usepackage{algorithmicx}
\usepackage{algpseudocode}

\definecolor{cvprblue}{rgb}{0.21,0.49,0.74}
\usepackage[pagebackref,breaklinks,colorlinks,allcolors=cvprblue]{hyperref}

\def\paperID{13220} 
\def\confName{CVPR}
\def\confYear{2025}

\title{MammAlps: A multi-view video behavior monitoring dataset of wild mammals in the Swiss Alps}

\author{
Valentin Gabeff\textsuperscript{1}\\
\and
Haozhe Qi\textsuperscript{1}\\
\and
Brendan Flaherty\textsuperscript{1}\\
\and
Gencer Sumbül\textsuperscript{1}\\
\and
Alexander Mathis\textsuperscript{1}\\
{\tt\small alexander.mathis@epfl.ch}
\and
Devis Tuia\textsuperscript{1}\\
{\tt\small devis.tuia@epfl.ch}
}

\affiliation{
\small
\textsuperscript{1} Ecole Polytechnique Fédérale de Lausanne (EPFL), Switzerland
}

\begin{document}

\twocolumn[{%
\maketitle

\begin{center}
    \centering
    \includegraphics[width=0.92\linewidth]{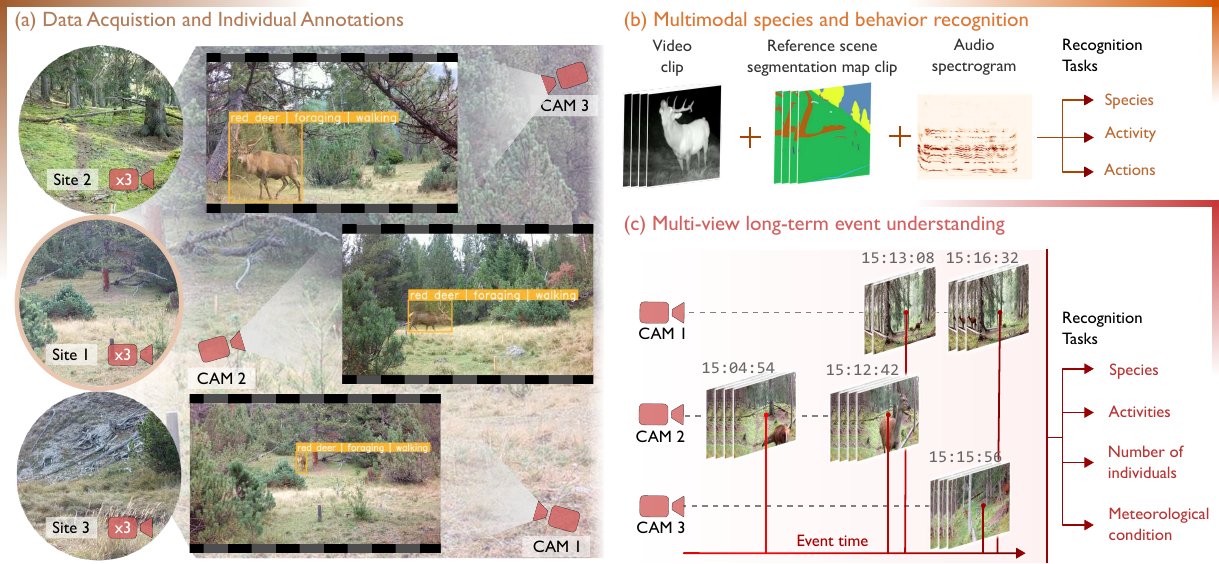}
    \captionof{figure}{\textbf{MammalAlps: Overview of the data and proposed benchmarks.} (a) Nine camera traps were installed at three different sites in the Swiss National Park and recorded video and audio of animal activity for six weeks. (b) We propose a multimodal species and hierarchical behavior recognition benchmark for wildlife based on video, audio and segmentation maps. (c) We propose the first multi-view, long-term event understanding benchmark that aims at summarizing long-term ecological events into meaningful information for behavioral ecology.}
    \label{fig:overview}
\end{center}
}]




\begin{abstract}
Monitoring wildlife is essential for ecology and ethology, especially in light of the increasing human impact on ecosystems. Camera traps have emerged as habitat-centric sensors enabling the study of wildlife populations at scale with minimal disturbance.
However, the lack of annotated video datasets limits the development of powerful video understanding models needed to process the vast amount of fieldwork data collected. To advance research in wild animal behavior monitoring we present MammAlps, a multimodal and multi-view dataset of wildlife behavior monitoring from 9 camera-traps in the Swiss National Park. MammAlps contains over 14 hours of video with audio, 2D segmentation maps and 8.5 hours of individual tracks densely labeled for species and behavior. Based on $6`135$ single animal clips, we propose the first hierarchical and multimodal animal behavior recognition benchmark using audio, video and reference scene segmentation maps as inputs. Furthermore,
we also propose a second ecology-oriented benchmark aiming at identifying activities, species, number of individuals and meteorological conditions from 397 multi-view and long-term ecological events, including false positive triggers. We advocate that both tasks are complementary and contribute to bridging the gap between machine learning and ecology. Code and data are available at \href{https://github.com/eceo-epfl/MammAlps}{https://github.com/eceo-epfl/MammAlps}. 
\end{abstract}

\section{Introduction}
\label{sec:intro}

Due to unprecedented rates of biodiversity loss, monitoring wild animals behavior has become a crucial task in conservation ecology and wildlife management~\cite{berger2011integrating, tobias2019integrating}. More broadly, understanding animal behavior is important across many fields~\cite{mathis2020deep,tuia2022perspectives,couzin2023emerging}. Wild animal behavior can be monitored with a variety of sensors. Animal-centric sensors such as bio-loggers are traditionally used to obtain broad behavioral information over large spatio-temporal extents~\cite{cooke2008biotelemetry, kays2015terrestrial,tuia2022perspectives,couzin2023emerging}. Conversely, habitat-centric imagery acquired from camera traps~\cite{caravaggi2017review, delisle2021next,tuia2022perspectives,couzin2023emerging} provides more fine-grained information on wildlife-environment interactions. With the most recent camera trap setups achieving enhanced battery life and storage, it is now becoming possible to study animal behavior at scale in the wild with video traps~\cite{brookes2024panaf20k, liu2024joint, liu2023lote}.

However, these advances in camera traps hardware also drastically increased dataset sizes, along with the complexity of the behavioral traits observed and to be quantified. To address this challenge, deep learning (DL) models were developed to support the analysis of wild animal videos for behavior recognition, segmentation and detection~\cite{beery2020context,schindler2021identification, fuchs2023asbar, brookes2023triple,  koger2023quantifying,schindler2024action,ye2024superanimal,brookes2024chimpvlm}. 

Simultaneously, wild animal datasets are being curated to support the training of DL models to effectively classify a wide range of behaviors across many species and geographical regions. Existing datasets annotated for wild animal behavior can generally be categorized in either fieldwork data, or internet scrapped data.
Fieldwork data is generally constrained to a small geographical location, focuses on one or few species and mostly contains common behaviors~\cite{tuia2022perspectives}. They have the advantage of representing “real world” data. In contrast, large scale datasets scrapped from the internet such as MammalNet~\cite{chen2023mammalnet} contain a rich set of behaviors and species, potentially with an over-representation of rare behaviors that are challenging to acquire in field surveys. Yet, they still suffer from an important domain gap between the videos scrapped (\eg scenes from documentaries) and the type of data used by experts (\eg camera trap imagery). Both sources of data are complementary, but the field still lacks publicly available and curated fieldwork datasets to unify them.
Additionally, insights from ethology and neuroscience can improve animal behavior recognition models by better representing behaviors in these wild animal datasets~\cite{anderson2014toward,stoffl2025elucidating}. Indeed, currently available datasets all categorize behaviors as independent classes, often without any kind of behavioral structure.

To address these shortcomings and advance research at the interface between computer vision and behavioral ecology, we collected and annotated MammAlps, a unique camera-trap video dataset consisting of footage acquired at three different sites in the Northern European Alps, at the Swiss National Park (SNP). MammAlps contains $8.5$ hours of curated mammals behavior recordings. Three cameras with varying level of field-of-view overlap were deployed at each site to provide multi-view information (\cref{fig:overview}a). Additionally, cameras built-in microphones were used to acquire audio and a segmentation map was created for each camera reference scene.
To better represent the hierarchical nature of animal behavior, individual tracklets were densely annotated at two levels of complexity, \ie high-level activities and low-level actions.

Along with the dataset, we propose the first multimodal species and behavior recognition benchmark from the camera trap video clips, the associated audio recordings and the reference scene segmentation map clips (\cref{fig:overview}b). 
We also provide a second benchmark consisting of summarized annotations at the event level (\eg a set of multiple videos capturing the same ecological scene) for long-term scene understanding task (\cref{fig:overview}c). This task consists of multiple predictive objectives at the event level from multiple views: Listing all detected species along with their activities, classifying the number of individuals into group sizes, and classifying meteorological conditions. In this second task, spatio-temporal precision is traded for larger spatio-temporal context which suits different needs in behavioral ecology. \\

\noindent Our contributions are:
\begin{itemize}
    \item A unique multimodal and multi-view camera-trap video dataset containing $8.5$ hours of densely annotated wild mammals behavior acquired in the Swiss Alps (\cref{fig:overview}a).
    \item A multimodal species and behavior recognition benchmark to foster method development for wildlife monitoring (\cref{fig:overview}b). 
    \item A unique multi-view and long-term event understanding benchmark designed to meet key unaddressed needs of ecologists, along with an offline method to condense long events into few visual tokens. (\cref{fig:overview}c).
\end{itemize}

\section{Related Works}
\label{sec:rel_works}

\begin{table*}[!t]
    \footnotesize
    \centering
    \begin{tabular}{l|c c c c c c c c c}
        \toprule
        \textbf{Dataset} & \textbf{\parbox{1.5cm}{\centering \textbf{Video hours} \\ \textbf{(processed)}}} & \textbf{Source} & \textbf{\# Videos} & \textbf{\# Species} & \textbf{\# Behav.} & \textbf{\parbox{1cm}{\centering \textbf{Annot.} \\ \textbf{level}}} & \textbf{\parbox{1cm}{\centering \textbf{Hierarch.} \\ \textbf{Behav.}}} &  \textbf{\parbox{1cm}{\centering \textbf{Multi-} \\ \textbf{Modal}}} & \textbf{\parbox{1cm}{\centering \textbf{Multi-} \\ \textbf{View}}} \\
        \midrule
        {Meerkats~\cite{rogers2023meerkat}} & 4 & Zoo & 35 & 1 & 15 & {individual} & {\xmark} & {\xmark} & {\xmark} \\
        {ChimpACT~\cite{ma2023chimpact}} & 2 & Zoo & 163 & 1 & 23 & {individual} & {\xmark} & {\cmark*} & {\xmark} \\
        {KABR~\cite{kholiavchenko2024kabr}} & 10 & Drone & 13k & 3 & 8 & {individual} & {\xmark} & {\xmark} & {\xmark} \\
        {BaboonLand~\cite{duporge2024baboonland}} & 20 & Drone & 30k & 1 & 12 & {individual} & {\xmark} & {\xmark} & {\xmark} \\
        {PanAf20k~\cite{brookes2024panaf20k}} & 80 & CT & 20k & 2 & 18 & {video} & {\xmark} & {\xmark} & {\xmark} \\
        {PanAf500~\cite{brookes2024panaf20k}} & 2 & CT & 500 & 2 & 9 & {individual} & {\xmark} & {\xmark} & {\xmark} \\
        {LoTE~\cite{liu2023lote}} & N/A & CT & 10k & 11 & 21 & {video} & {\xmark} & {\cmark*} & {\xmark} \\
        {PandaFormer~\cite{liu2024joint}} & 2 & CT & 1431 & 1 & 5 & {video} & {\xmark} & {\xmark} & {\xmark} \\
        \midrule
        {AnimalKingdom~\cite{ng2022animal}} & 50 & Youtube & 30k & 850 & 140 & {video} & {\xmark} & {\cmark*} & {\xmark} \\
        {MammalNet~\cite{chen2023mammalnet}} & 394 & Youtube & 20k & 173 & 12 & {video} & {\xmark} & {\xmark} & {\xmark} \\
        \midrule
        MammAlps (clips) & 8.5 & CT & 6k & 5 & 11+19 & {individual} & {\cmark} & {\cmark} & {\xmark} \\
        MammAlps (events) & 14.5 & CT & 2384 & 5 & 11 & {event} & {\cmark} & {\cmark*} & {\cmark} \\
        \bottomrule
    \end{tabular}
    \caption{\textbf{Prominent and publicly available video datasets of wild animals behavior monitoring.} *Multimodal data is available but it is not used for an action recognition benchmark. MammAlps is available at \href{https://doi.org/10.5281/zenodo.15040900}{10.5281/zenodo.15040900}.}    
    \label{tab:rel_works}
\end{table*}

\textbf{Wild animal behavioral datasets.} Thanks to advances in sensor design and availability~\cite{tuia2022perspectives,couzin2023emerging}, a number of fieldwork-based datasets for wildlife behavior monitoring from videos became available recently (\cref{tab:rel_works}). LoTE offers a collection of camera trap datasets (images and videos) from South East Asia~\cite{liu2023lote}. While a subset of the images are labeled with bounding boxes, the behavior annotations for the video dataset are not at the individual level. Brookes et al. share a camera trap video dataset of great apes in Africa~\cite{brookes2024panaf20k}. Each video is associated with a set of behavior labels that occur within the video, and a subset of the dataset also comprises individual tracks. A larger part of the dataset contains richer behavior descriptions, yet without individual tracks. 
The meerkat behavior dataset contains rich behavioral annotations at the individual level~\cite{rogers2023meerkat}. Similarly, ChimpACT contains individual level annotations, along with animal body pose annotations~\cite{ma2023chimpact}. However, both datasets are recorded in zoos. 
KABR and BaboonLand use drone footage and provide dense behavior labels for four African species at the individual level~\cite{duporge2024baboonland,kholiavchenko2024kabr}.
PandaFormer~\cite{liu2024joint} contains almost two hours of wild pandas recordings spanning five behaviors. Recently, a 1-h long dataset  with recordings of 17 bird species and seven behavioral classes became available\cite{rodriguez2025visual}.

Scraping the web can also yield relevant datasets. Animal Kingdom~\cite{ng2022animal} contains $50$ hours of behavioral videos spanning $850$ species and $140$ behavioral descriptions. MammalNet~\cite{chen2023mammalnet} is the largest dataset of wild animal videos, containing around $400$ hours of footage from different sources (\eg documentaries, zoos) depicting $173$ mammal species and around $20$ behaviors shared across mammals.
While some of these works propose exclusively low-level behavior recognition~\cite{kholiavchenko2024kabr, liu2024joint} (\eg actions like walking, grazing), others annotate more high-level behaviors~\cite{chen2023mammalnet, brookes2024panaf20k, duporge2024baboonland} (\eg chasing, hunting). \\
\textbf{Multi-modal action recognition.} With the development of the transformer architecture \cite{vaswani2017attention} and expanding computational power, leveraging multimodal data for action understanding was increasingly feasible\cite{shah2023multi, wang2019generative, zhao2023learning, chalk2024tim, xiao2020audiovisual, shamil2024utility}. LaViLa \citep{zhao2023learning} learns video representations from pre-trained large language models. TIM~\citep{chalk2024tim} designs time interval encodings to incorporate visual and audio events. In the domain of wildlife behavior understanding, researchers sometimes use multiple sensors (\ie modalities) conjointly to monitor animal behavior~\cite{alempijevic2021natural, handley2018behaviourally, bain2021automated}. In ~\cite{bain2021automated}, the authors make a first attempt at using audio-visual inputs from camera traps to classify two specific wild primate behaviors.

Overall, our work is most similar to~\cite{kholiavchenko2024kabr,duporge2024baboonland, brookes2024panaf20k, bain2021automated}. On top of the dense behavioral annotations at the individual level, our dataset brings additional value over all previous datasets as (1) we follow a hierarchical representation of behavior~\cite{anderson2014toward,stoffl2025elucidating}, and propose separate tasks for low-level action and high-level activity recognition; (2) we provide audio recordings and segmentation maps from the fixed camera reference scenes to further guide models via multiple modalities; (3) events are being recorded from up to three points-of-view, which provides detailed information for long-term event understanding (\cref{tab:rel_works}); (4) MammAlps is the only camera trap video dataset focusing on species from the European Alps, which is a region particularly vulnerable to climate change~\cite{vitasse2021phenological, gobiet201421st}.

\section{MammAlps dataset and proposed benchmarks}
\label{sec:mammalps}

\begin{figure*}
    \centering
    \includegraphics[width=1.0\linewidth]{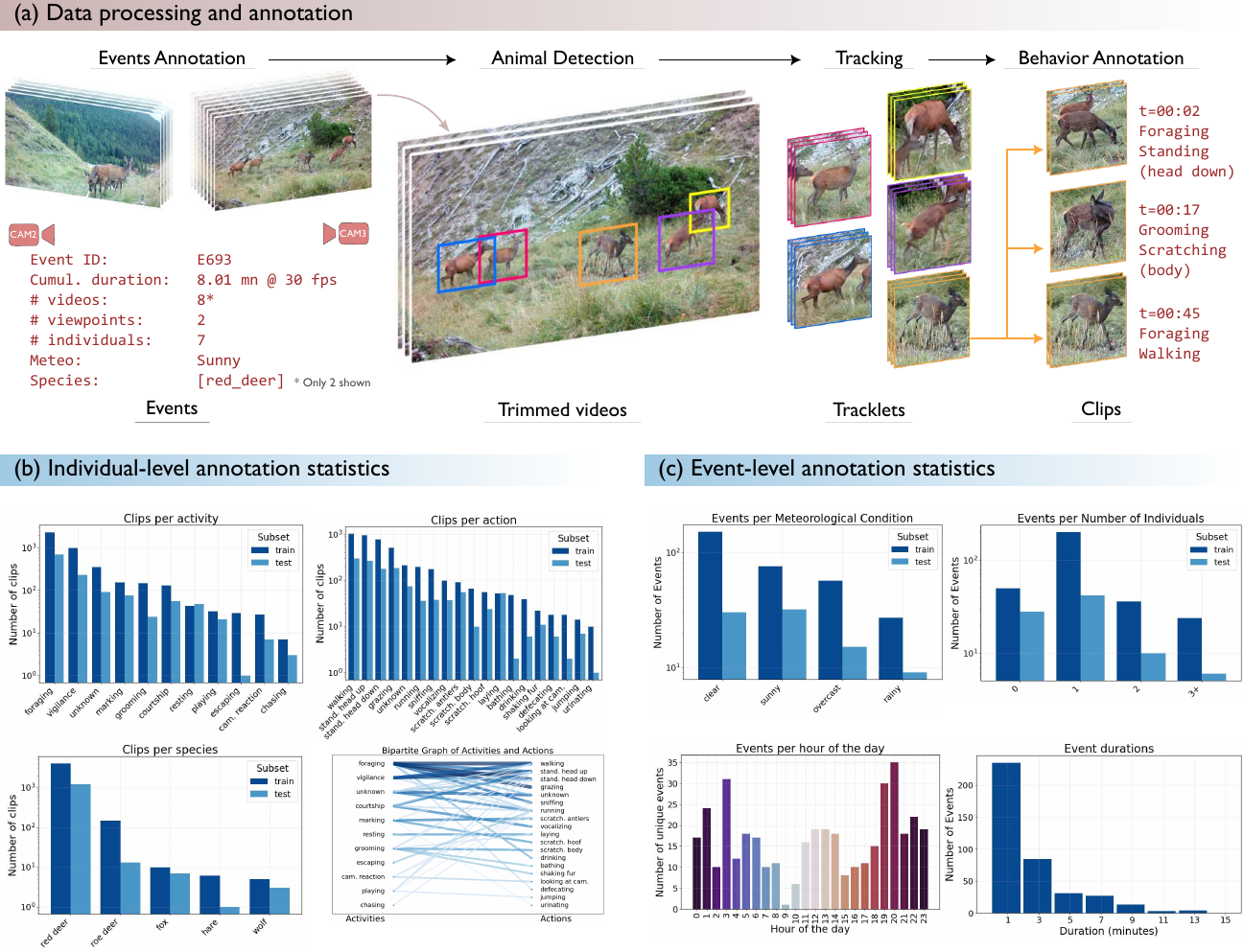}
    \caption{\textbf{Data processing pipeline and analysis.} (a) Raw videos were first aggregated into events. We then applied MegaDetector~\cite{beery2019efficient, hernandez2024pytorchwildlife} and ByteTrack~\cite{zhang2022bytetrack} to generate animal tracklets, which were manually corrected. We annotated these tracklets for species and behavior at two levels of complexity. (b-c) Various statistics of the dataset.}
    \label{fig:mammalps}
\end{figure*}

\begin{table}[]
    \begin{tabular}{l p{.35\textwidth}}
    \toprule
        \makecell[tl]{Raw \\ video} & Raw camera trap recording of fixed duration. \\\midrule
        \addlinespace[0.5ex]
        Event & Collection of raw videos corresponding to an ecological event. Events are separated by a period of inactivity of at least 5 minutes. The events are used as input for the long-term scene understanding task (\cref{sec:mammalps-b2}). \\\midrule
        \addlinespace[0.5ex]
        \makecell[tl]{Trimmed \\ video} & Segment within a raw video contained between the first and the last MegaDetector~\cite{beery2019efficient, hernandez2024pytorchwildlife} detections. \\\midrule
        \addlinespace[0.5ex]
        Track & Sequence of bounding boxes with associated individual identifier, built automatically from ByteTrack~\cite{zhang2022bytetrack} and MegaDetector predictions~\cite{beery2019efficient, hernandez2024pytorchwildlife} and manually adjusted in CVAT~\cite{CVAT_ai_Corporation_Computer_Vision_Annotation_2023}.\\\midrule
        \addlinespace[0.5ex]
        Tracklet & Animal-centered video of aspect ratio 1:1 cropped from an animal track labeled for species and densely annotated for behavior. \\\midrule
        \addlinespace[0.5ex]
        Clip & Segment within a tracklet with a single behavioral expression. The clips are used as input for the behavior recognition benchmark (\cref{sec:mammalps-b1}). \\
    \bottomrule
    \end{tabular}
    \caption{\textbf{Terminology used at the different stages of the data processing and annotation pipeline.}}
    \label{tab:terminology}
\end{table}

In this section, we detail the dataset collection and pre-processing (\cref{sec:mammalps-data}) of MammAlps, as well as the annotation protocol (\cref{sec:mammalps-annot}) and the two benchmarks proposed (\cref{sec:mammalps-b1} and \ref{sec:mammalps-b2}). For clarity, we defined a list of terms used throughout the study in~\cref{tab:terminology}.

\subsection{Data collection and pre-processing}
\label{sec:mammalps-data}
\textbf{Data collection.} Nine camera traps (Browning's Spec Ops Elite HP5) were installed in the Swiss National at three sampling sites representing different ecological habitats. The project was approved by the Research Commission of the National Park. For each site, three cameras were positioned with different perspectives, in order to capture the scene from multiple angles and to provide more context for interpreting behavior (\cref{fig:overview}a). Triggered by motion, videos were collected for six weeks (between June and August 2023) during daytime and nighttime. At nighttime, videos are recorded with an IR flash invisible to the species of interest. Videos are captured at high resolution ($1920\times1080$) with a frame rate of $30$ FPS. Cameras recorded $43$ hours of raw footage, with varying levels of false positive triggers. Data acquisition details and sampling site descriptions can be found in the Supplementary Materials. \\
\textbf{Data pre-processing.} The data processing pipeline is as follows (\cref{fig:mammalps}a): \textit{raw videos} were first grouped into \textit{events}, corresponding to periods without more than five minutes of inactivity at the corresponding site. We then removed false positive videos and trimmed the true positive ones by running them through MegaDetector~\cite{beery2019efficient, hernandez2024pytorchwildlife}. The dense animal detection predictions of the \textit{trimmed video} were used as inputs to an adapted version of ByteTrack~\cite{zhang2022bytetrack} yielding individual \textit{tracks}. The \textit{tracks} were then manually corrected in CVAT~\cite{CVAT_ai_Corporation_Computer_Vision_Annotation_2023} to remove identity switches, lost tracks, and any remaining false positive segment. We did not correct localization errors (\eg body parts outside of bounding boxes) since our proposed benchmarks do not require this level of spatial precision. Each animal track was converted into a video \textit{tracklet} ($380\times380$) padded with background to avoid distortions. We further partition the tracklets into short video \textit{clips} displaying a single behavioral expression (\cref{sec:mammalps-annot}). Data processing and model parameters are detailed in the Supplementary Materials. \\
\textbf{Cameras synchronization and temporal drift.} Cameras have a built-in accuracy of one minute and are subject to drift over time (see Supplementary Materials). Temporal drift between camera pairs extended up to one minute in Site 1. This drift further increases the difficulty of the second benchmark, while reflecting fieldwork conditions. Auditory data could be used for syncing. 

\subsection{Data annotation}
\label{sec:mammalps-annot}

Individual counts and meteorological conditions were annotated at the event level, while behaviors and species were annotated at the individual level (\cref{fig:mammalps}b) and then aggregated at the event level when necessary (\cref{fig:mammalps}c).\\
\textbf{Species and behavior annotations.} Animal \textit{tracklets} were densely labeled in CVAT~\cite{CVAT_ai_Corporation_Computer_Vision_Annotation_2023} for species and behaviors. We focused on five species: red deer (\textit{Cervus elaphus}), roe deer (\textit{Capreolus capreolus}), fox (\textit{Vulpes vulpes}), wolf (\textit{Canis lupus}) and mountain hare (\textit{Lepus timidus}). 
Behaviors were annotated by two annotators at two levels of complexity~\cite{anderson2014toward,stoffl2025elucidating}: 1) \textit{Actions} (\eg walking, grazing), are stereotypical combinations of a few basic movements and can usually be identified from a few frames; 2) \textit{Activities}, which generally require longer spatio-temporal context and may be the composition of multiple \textit{actions} (\eg foraging) or the interaction between different individuals of the same species (\eg courtship) or between different species (\eg chasing). Each frame is labeled with one activity and either one or two non-mutually exclusive actions.
For both levels, we included an \textit{unknown} class, which indicates a behavior that could not be identified, either because of occlusion or by lack of information. \\
\textbf{Individual counts.} The total number of individuals in an event is determined by visual examination of all the videos from all viewpoints recording it. Automatic aggregation from the track annotations was not possible, since camera traps were not perfectly temporally synchronized nor spatially referenced in a 3D model. Individual counts per species were summed and grouped into four categories (0, 1, 2, 3$+$). Thus, the counting task assesses the group size (none, individual, pair or group).\\
\textbf{Meteorological conditions.} During this process, meteorological conditions were visually determined and categorized into four general conditions: \emph{clear weather} (including day and night), \emph{sunny}, \emph{overcast} and \emph{rainy} (including day and night). \\
\textbf{Reference scenes segmentation.} Since camera traps are placed at a fixed position, a single segmentation map was annotated for each of the scene's viewpoints. A reference picture (without animals) was taken with each camera after the video acquisition. We annotated the segmentation masks for ten classes using CVAT~\cite{CVAT_ai_Corporation_Computer_Vision_Annotation_2023}. Some classes are unique to a site (\eg water pound only occurs at Site 3), while others are shared across the three sites (\eg grass). The segmentation maps are then processed into video clips by generating a tracklet based on the animal tracks for every video clip. Hence, these segmentation map clips represent the background classes surrounding (and behind) the animal, synchronized in location and time to the animal video clips. Examples are shown in the Supplementary Materials.

\subsection{Multimodal Species and Behavior Recognition Benchmark: B1}
\label{sec:mammalps-b1}

Action recognition is a common challenge across multiple wildlife monitoring datasets~\cite{kholiavchenko2024kabr, duporge2024baboonland, brookes2024panaf20k, liu2024joint, chen2023mammalnet, ng2022animal}. While all of them are limited to RGB visual inputs, we enrich the video modality with audio and reference scene segmentation maps. We hypothesized that audio can help identify some specific actions such as vocalization and walking, while segmentation maps of the reference scenes may guide classification for behaviors involving specific environmental features (\eg drinking from a water source).
The dataset for this task (B1) consists of $6135$ short video clips spanning $11$ activities, $19$ actions and $5$ species and a total of $8.5$ hours of recordings. Because a sample can be annotated with up to two actions, action recognition is a multi-label classification task, while species and activity recognitions are multi-class ones. We refer to \textit{behavior} recognition as the recognition task that encompasses both action and activity recognition. The data was randomly split in a train, validation and test set at the day level, while matching label distributions across splits. Clips that contained occlusions were labeled as unknown activity and actions since we considered that a model cannot provide a reliable behavioral estimate with such limited context.

\begin{table}[t]
    \centering
    \begin{tabular}{l c c c}
    \toprule
        Training task & Spe.($\uparrow$) & ActY.($\uparrow$) & ActN.($\uparrow$) \\
    \midrule
        \multicolumn{4}{c}{Single Task Prediction}  \\
    \midrule
         Spe. & 0.537 & - & -  \\
         ActY. & - & 0.440 & -  \\
         ActN. & - & - & 0.447  \\
    \midrule
        \multicolumn{4}{c}{Joint Task Prediction}  \\
    \midrule
        Spe. + ActY. & 0.437 & \textbf{0.443} & - \\ 
         Spe. + ActN. & \textbf{0.539} & - & 0.442 \\
         ActY. + ActN. & - & 0.442 & 0.427 \\
         All. & 0.487 & 0.428 & \textbf{0.458} \\
    \bottomrule
    \end{tabular}
    \caption{{\bf Comparison of single vs. joint task prediction (B1).} mAP for single and joint task predictions from video clips. In all cases, VideoMAE is used as the base model~\cite{tong2022videomae}. ActY.: Activities; ActN.: Actions; Spe.: Species.}
    \label{tab:results_b1_unimodal}
\end{table}

\subsection{Multi-view long-term event understanding Benchmark: B2}
\label{sec:mammalps-b2}

Benchmark B1 is a computer science-oriented benchmark focused on a single sensor (with multiple modalities). However, to reliably identify events all the available sensors should be used. Additionally, understanding events requires long-term context understanding (more than $16$ frames), especially when expressed activities are temporally related to other individuals (\eg prey-predator relationships) or are composed of multiple actions (\eg foraging). 
Being able to efficiently summarize events into broad categories is also necessary to facilitate the annotation process of very large camera trap datasets. To this end, we propose a second, long-term event understanding benchmark (B2) that takes as input the raw multi-view videos of a given event with the objective of predicting high-level behaviors (activities), the species detected, the number of individuals (in the grouped categories defined in~\cref{sec:mammalps-annot}) and the meteorological conditions. Activity and species recognition are multi-label classification tasks, while meteorological condition and number of individuals are multi-class classification ones. This benchmark is particularly challenging as the event duration varies greatly (from $1$ second to $12$ minutes), activities and species are highly imbalanced, and counting individuals requires to intelligently integrate information across camera views and over time.
The dataset for task B2 is composed of $397$ events, $2384$ videos, totaling $14.2$ hours of recordings. Similarly as for~\cref{sec:mammalps-b1}, the events were randomly split (at the day level) in a train and test set. Data spans $11$ activities and $5$ species (the same as for~\cref{sec:mammalps-b1}), $4$ group size categories and $4$ meteorological conditions.

\begin{table}[t]
    \centering
    \begin{tabular}{ l c c c c}
    \toprule
        Mod. & Spe.($\uparrow$) & ActY.($\uparrow$) & ActN.($\uparrow$) & Avg.\scriptsize{$\pm$std} \normalsize{($\uparrow$)} \\
    \midrule
        V & 0.495 & 0.436 & 0.452 & 0.453 \scriptsize{$\pm$ 0.002} \\ 
        S & 0.441 & 0.234 & 0.172 & 0.230 \scriptsize{$\pm$ 0.014} \\ 
        A & 0.223 & 0.212 & 0.172 & 0.192 \scriptsize{$\pm$ 0.004} \\ 
        V+S & 0.466 & 0.409 & 0.384 & 0.403 \scriptsize{$\pm$ 0.006} \\ 
        A+S & 0.385 & 0.312 & 0.276 & 0.303 \scriptsize{$\pm$ 0.012} \\ 
        V+A & 0.473 & 0.484 & \textbf{0.466} & \textbf{0.473} \scriptsize{$\pm$ 0.017} \\ 
        V+A+S & \textbf{0.531} & \textbf{0.485} & 0.437 & 0.466 \scriptsize{$\pm$ 0.007} \\ 
    \bottomrule
    \end{tabular}
    \caption{{\bf Hierarchical action recognition from multimodal data (B1).} mAP for joint task prediction from multimodal data using VideoMAE as the base model~\cite{tong2022videomae} averaged across three runs. Mod.: Modalities; V: video clips; A: audio spectrograms; S: segmentation map clips;  ActY.: Activities; ActN.: Actions; Spe.: Species; Avg.: overall per-class average.}
    \label{tab:results_b1_multimodal}
\end{table}

\begin{table*}[t]
    \centering
    \begin{tabular}{l c c c c c c c}
    \toprule
        Training task & $r$ & Cont. Len. & Spe.($\uparrow$) & ActY.($\uparrow$) & Met. Cond.($\uparrow$) & Indiv.($\uparrow$) & Avg.\scriptsize{$\pm$std} \normalsize{($\uparrow$)} \\
    \midrule
        \multicolumn{8}{c}{Single Task Prediction}  \\
    \midrule
        Spe. & 14 & 4096 & \textbf{0.481} & - & - & - &  \\
        ActY. & 14 & 4096 & - & 0.478 & - & - &  \\
        Met. Cond. & 14 & 4096 & - & - & \textbf{0.681} & - &  \\
        Indiv. & 14 & 4096 & - & - & - & \textbf{0.592} &  \\
    \midrule
        \multicolumn{8}{c}{Joint Task Prediction} \\
    \midrule
        All & 14 & 4096 & 0.415 & 0.479 & 0.618 & 0.499 & 0.489 \scriptsize{$\pm$ 0.033} \\ 
        All & 11 & 8192 & 0.446 & \textbf{0.481} & 0.594 & 0.543 & \textbf{0.500}  \scriptsize{$\pm$ 0.004} \\ 
    \bottomrule
\end{tabular}

    \caption{ {\bf mAP for long-term event understanding from the multi-view events (B2).} Results are averaged across three runs for the joint task predictions. All models use the transformer encoder from ViT-Base. "$r$": ToME~\cite{bolya2022token} reduction factor. A larger reduction factor leads to more patches being merged at the frame level and fewer video tokens; "Cont. Len.": context length: number of tokens per sample; ActY.: Activities; Spe.: Species.; Met. Cond.: Meteorological Conditions; Indiv.: Number of individuals categories.; Avg.: overall per-class average.}
    \label{tab:results_b2}
\end{table*}

For both benchmarks B1 and B2, we report the mean average precision (mAP) averaged over the label categories of each task, which is a convenient metric to compare tasks that are either multi-label or multi-class. When applicable, for joint predictions on all tasks, we report the mAP averaged over all label categories of all tasks in column `Avg.'. For the joint task predictions, we report the mAP  averaged across three runs using different seeds (along with the std. for the `Avg.` column). Models for benchmarks B1 and B2 were trained with four and eight A100 GPUs, respectively.

\section{Experiments}
\label{sec:results}

\subsection{B1: Multi-modal species and behavior recognition}

In order to utilize multi-modal data for action recognition, we adapted the VideoMAE model \cite{tong2022videomae} so that it could take video, audio and segmentation maps as inputs simultaneously. Specifically, we sampled $16$ frames within $5$ seconds of randomly selected windows for both video and one-hot encoded segmentation map clips. For the audio inputs, we first found the audio clip simultaneous to the video clip and then transformed and tokenized the original audio signal to a spectrogram, similarly to AudioMAE~\cite{huang2022masked}. 
To compensate for the label imbalance, clips were sampled with a probability proportional to the sum of the inverse label frequencies for each class. Because test clips greatly vary in their duration, we aggregated predictions over ten random samples of $16$ frames for every test clip.

When considering only the video modality, VideoMAE leads to improved results for all tasks when considering the joint task prediction (\cref{tab:results_b1_unimodal}). Multi-modal results indicate that combining the audio and video modalities improves the performance over the video-only model ($+0.020$ mAP), with an overall class-average mAP of $0.473$ (\cref{tab:results_b1_multimodal}). However, in our baseline model, the reference segmentation map clips did not improve over their video-only or video-audio counterparts, but they did increase the performance of the audio-only model ($+0.111$ mAP) suggesting that this modality contains distinct information relevant to the tasks. More details, baselines and results per class can be found in the Supplementary Materials. 

\subsection{B2: Multi-view long-term event understanding}

To the best of our knowledge, due to the size no existing video model can process multi-view and long-term (ecological) data for our task of interest, so we propose a simple method as baseline. Taking inspiration from token merging in vision transformers (ToME)~\cite{bolya2022token} and follow-up works focusing on merging tokens online over time~\cite{shen2024tempme, ren2023testa}, we propose a fully offline method to merge the frame patch tokens from a pretrained vision-MAE transformer first in the spatial and then in the temporal dimensions (see Supplementary Materials). To account for the large range of video durations, we perform token merging over time in blocks of fixed duration and concatenate the resulting tokens, so that long videos ultimately yield more tokens than short ones. We add three cosine positional embeddings~\cite{dosovitskiy2020image} to every video token: 1) The information from the camera identity for the given site (Cam$_{ID}$); 2) the elapsed time with respect to the event start ($\Delta T_{event}$); and 3) the frame and patch identities of the source frame tokens composing each individual video token (see Supplementary Materials for details). We input these condensed video tokens to a transformer backbone with four output heads, each corresponding to one of the predictive tasks. We set a maximum input context length based on the longest event and pad shorter ones with masked tokens. 

The best joint recognition performance (average per-class mAP of $0.500$) was achieved with a ToME~\cite{bolya2022token} reduction factor ($r$) at the frame level of 11, yielding between $65$ and $390$ tokens per video depending on their duration (\cref{tab:results_b2}). When $r=11$, the overall mAP is slightly higher ($+0.011$) than when $r=14$ (yielding between $29$ and $174$ tokens per video) but not on all tasks. 

We evaluated the model performance when ablating $r$ and the different positional embeddings (\cref{tab:ablation_b2}). We focused on the task where these embeddings are thought to contribute the most: number of individuals classification. Here, the model with all positional embeddings lead to the highest scores independent of the value of $r$.  While with $r=14$, the highest increase is observed for the single task ($+0.078$ mAP), this is the opposite when $r=11$ (increase in joint task mAP of $+0.109$).  More details, ablations and results per class can be found in the Supplementary Materials. 

\begin{table*}[t]
    \centering
    \begin{tabular}{c c | c c c | c c }
    \toprule
        \multicolumn{2}{c|}{ToME parameters} & \multicolumn{3}{c|}{Positional embeddings} & \multicolumn{2}{c}{mAP} \\
         r & Cont. Len./BS & Cam$_{ID}$ & $\Delta T_{event}$ & Source & \parbox{2cm}{\centering Indiv.($\uparrow$) \\ (Single$|$Joint)} & \parbox{2.5cm}{\centering Indiv. 2+ ($\uparrow$) \\ (Single$|$Joint)} \\
    \midrule
          14 & 4096/32 & & & & 0.514$|$0.505 & 0.222$|$0.120 \\
          14 & 4096/32 & \cmark & \cmark & & 0.562$|$0.461 & 0.192$|$0.112 \\
          14 & 4096/32 & \cmark & \cmark & \cmark & \textbf{0.592}$|$0.478 & \textbf{0.329}$|$0.156 \\
    \midrule
          11 & 8192/8 &  &  & & 0.502$|$0.566 & 0.145$|$0.223 \\
          11 & 8192/8 & \cmark & \cmark & & 0.527$|$0.484 & 0.200$|$0.136 \\
          11 & 8192/8 & \cmark & \cmark & \cmark & 0.527$|$\textbf{0.593} & 0.184$|$\textbf{0.294} \\
    \bottomrule
    \end{tabular}
    \caption{ {\bf Ablation study on the effect of the number of video tokens, and the addition of the different positional embeddings on the number of individuals recognition task (B2).} All models are the transformer encoder from ViT-Base. "r": ToME~\cite{bolya2022token} reduction factor."Cont. Len.": context length: maximum number of tokens per sample; BS: Batch Size; ActY.: Activities; Spe.: Species; Met. Cond.: Meteorological Conditions; Indiv.: Number of individuals in categories; `Indiv. 2+': Predictions for groups containing more than a single individual. Results for the `Indiv.' and `Indiv. 2+' tasks are provided for both the single and joint task prediction.}
    \label{tab:ablation_b2}
\end{table*}

\section{Discussion}
\label{sec:discussion}

\textbf{Contributions of the audio and segmentation map modalities.}
Adding the audio modality improves the overall performance over a video-only model (\cref{tab:results_b1_multimodal}).
When looking at specific classes (Supplementary Materials), classes with distinct sounds such as marking or vocalizing improved for the audio-video model over the video-only model ($+0.20$ and $+0.09$ F1-scores, respectively). Conversely, the resting activity which is mostly silent remains with a low F1-score (from $0.19$ with video to $0.15$ with the audio and video). While the reference segmentation map modality did not improve performance when combined with videos, it did improve over the audio-only model especially on classes involving specific scene features such as grazing ($+0.08$) or walking ($+0.09)$ despite that these actions already emit some sound. \\
\textbf{Impact of token merging on classifying the number of individuals.} In B2 (\cref{sec:mammalps-b2}), classifying the number of individuals is particularly challenging as the model needs to integrate information from multiple views and videos. Hence the model needs to extract individual identities. Yet, it is common that tokens representing different animals become merged by our offline approach. This is expected as the algorithm merges tokens based on similarity and two different individuals might show little visual differences when they are from the same species. We address this issue by both increasing the number of tokens per video and by adding a positional embedding to the video tokens that contains summarized information about their source frames and patches. With the former, we aim that different individuals are represented by different tokens, while with the latter, we indicate if a single video token comes from one or multiple discontinuous spatio-temporal segments. The ablation realized suggests that this design successfully increases the performance for test events with more than two individuals (\cref{tab:ablation_b2}). \\
\textbf{Hierarchical description of behaviors.} The decision to represent behaviors as a combination of one activity and one or two actions seems to facilitate learning, as suggested by the higher performance of joint recognition models over single ones (\cref{tab:results_b1_unimodal}). 
However, the hierarchical relationship between activities and actions could be further exploited in both benchmarks. For example in B2 (\cref{sec:mammalps-b2}), predicted actions could influence higher-level activity prediction, \eg chasing is a high-level activity composed of the running action in a prey-predator context.\\
\textbf{Dataset bias and limitations.} 
First, annotating animal behavior is a complex task, as behavior categorization remains a subjective process, prone to annotators' biases. This concerns particularly social behaviors such as those related to courtship. These uncertainties lead to varying level of label noise per class. To mitigate these biases, uncertain samples were tagged and discussed among annotators to ensure annotation consistency.  
Additionally, the set of species in the dataset remains limited, as all three sites were located in the same National Park, at the same elevation and over a short period of time (\ie until the camera battery exhaustion). This limits the possibility to learn common behavioral expression across species as done in~\cite{chen2023mammalnet}. Other mammal species that are common in the European Alps are absent from the dataset in its current form.
Likewise, despite containing $80$GB of raw video data, the dataset of the long-term video understanding benchmark only contains $86$ test events, which may be insufficient to properly assess the performance of the model on rare classes. However, this is the first dataset considering events-level information in the wildlife domain and which defines a new task for the field. Future surveys (by the authors themselves and desirably by the wider and very active `AI for ecology' community) will progressively increase the quantity of the data for this task.

\section{Conclusion and outlook}
\label{sec:conclusion}

We develop MammalAlps, a novel multimodal, multi-view camera-trap video dataset of annotated hierarchal, mammalian behavior in the Swiss Alps. We propose two benchmarks to motivate the development of behavior understanding methods for ecology, based on event and clip annotations. In particular, we propose the first long-term event understanding task that aims to summarize long-term ecological events into meaningful information for the ecologists. 
We believe this task is particularly interesting to spur research on efficient architectures that can flexibly integrate multiple sources of information over diverse temporal ranges to reach better conclusions. 
 
MammAlps can be extended in a multitude of ways, for instance by adding new modalities such as animal body pose~\cite{ye2024superanimal}, segmentation masks~\cite{ravi2024sam}, depth~\cite{yang2024depth}, and language~\cite{gabeff2024wildclip, brookes2024chimpvlm}, as all these modalities introduce complimentary behavior information.

By publicly sharing MammAlps, we aim to provide rich behavioral annotations that can fuel the development of holistic animal behavior understanding models. These models have the potential to identify and quantify observable behavioral traits of wild individuals, opening the doors to AI-assisted data processing and scientific discoveries.

{\bf Acknowledgments:} We thank members of the Mathis Group for Computational Neuroscience \& AI (EPFL) and of the Environmental Computational Science and Earth Observation Laboratory (EPFL) for their feedback and fieldwork efforts. We also thank members of the Swiss National Park monitoring team for their support and feedback. This project was partially funded by EPFL's SV-ENAC I-PhD program (G.V.), Boehringer Ingelheim Fonds PhD stipend (H.Q.) and Swiss SNF grant (320030-227871).

\vspace{1em} 
{\noindent {\bf Changes from CVPR published version and arXiv v1:} After submission, we noticed that a few audio files were not correctly aligned with the corresponding video. We fixed the issue, which had little to no impact on performance (\cref{tab:results_b1_multimodal}). The released data on Zenodo already reflects this correction: \href{https://doi.org/10.5281/zenodo.15040900}{10.5281/zenodo.15040900}. During this process, we also ran more replicates. Consequently, results for \cref{tab:results_b1_multimodal} and for the joint task predictions of \cref{tab:results_b2} have been updated to show results for three runs. We now report the mAP averaged across three runs along with the standard deviation for the average mAP to provide a better estimate of the performance variation across runs. We report the standard deviation for all tasks in the Supplementary Material. Results in the text have been updated accordingly.}

{
    \small
    \bibliographystyle{ieeenat_fullname}
    \bibliography{main}
}

\maketitlesupplementary

\tableofcontents


\section{Data acquisition}

\subsection{Site descriptions and period of acquisition}

The Swiss National Park is located in Eastern Switzerland and has a substantially higher density of ungulates compared to neighboring regions. Additionally, the park is a strictly protected nature reserve, and thus human activities are restricted to be minimal~\cite{fluri2023influence}. This makes the region particularly interesting to acquire data on the naturalistic behaviors of ungulates from camera trap videos over a relatively short period of time. 

We identified three sites for habitat monitoring. The three sites used for the study are located between 1840 m and 1890 m of altitude, at which elevation mostly red deers and roe deers are found, chamois foraging generally higher at this period of the year. For privacy reasons, we do not disclose the exact location.

Site 1 is a clearing within an alpine forest composed of larch, cembra pine, mountain pine and spruce facing South-West. Site 2 is located at the intersection of multiple game paths, in a similar forest type facing North. Site 3 is located by a water stream where the terrain creates two small water pounds, and is facing towards South. The three sites were chosen by purpose to acquire a behavioral dataset as diverse as possible since observing different behavioral expressions is of a high chance in these sites.
Cameras acquired video and audio data for 6 weeks between August and October 2023. This period corresponds to the rutting season of red deer, and thus many events represent rutting-related behaviors.


\subsection{Camera settings}

Camera traps (Browning's Spec Ops Elite HP5) acquired videos of fixed duration (either for 1 or 2 minutes at daytime, and 20 seconds when with the IR flash). Cameras were set to fast trigger mode with a delay of 1 second between subsequent videos, with long-range motion detection enabled.
Cameras were fixed either on wooden poles or on trees, around 60 cm above ground. Cameras were positioned on the sites with varying levels of field-of-view overlap, while Site 3 had the most con-focal setup, and site 1 had the least.

We report video acquisition statistics per camera and per site (~\cref{fig:acquisition_stats}a-b). When cameras began to run out of battery, the recordings at night were automatically shortened by the hardware, leading to many nighttime clips with short durations (below 20 seconds). Among these clips, We kept only the ones containing at least 30 frames (1 second).

\section{Details on data processing and annotation}
\subsection{From events to tracklets}

We used MegaDetector~\cite{beery2019efficient, hernandez2024pytorchwildlife} v5a at a sampling period of five frames to detect recordings with animals among the raw videos ($N=3794$). Videos which did not have at least two animal detections above a permissive animal detection confidence threshold of $0.3$, were considered as false positives. The videos with detections ($N=1961$) were then trimmed to the segment between the first and last MegaDetector detection. We ran MegaDetector v5a again on every frame of the trimmed videos to obtain dense animal detection predictions.

To obtain animal tracks we adapted the matching algorithm from ByteTrack~\cite{zhang2022bytetrack}. Indeed, ByteTrack performance depends on the performance of the object detector and the frame rate (the more frequent the better). However, as MegaDetector was not fine-tuned on our data, we observe a high rate of missing detections either because of long-term occlusions (\eg an animal passing behind a tree), low frame quality (\eg at night), and relatively low frame rate (for tracking, i.e., $30$ FPS).
To improve tracking performance, we used the generalized intersection-over-union matching cost (GIoU), instead of the (IoU) originally proposed in ByteTrack to allow the matching of bounding boxes even when they do not overlap. We added an area difference matching cost to avoid matching animals with small false detections from MegaDetector (\eg rain drops). We also gave maximum certainty to the measurements (MegaDetector bounding boxes) during the Kalman Filter integration process to avoid long-term interpolations and bounding boxes that would lag behind the animal after long occlusions. Specifically, we used a detection threshold of $0.2$, a track activation threshold of $0.5$, a lost track buffer of $300$ frames, and a minimum matching threshold for high confidence pairs of $0.75$. The cost $C$ between two bounding boxes $B_i$ and $B_j$ is defined as follows:
\begin{equation}
\begin{split}
    C(B_i, B_j) & = 1 - (GIoU(B_i, B_j) - 2*A(B_i, B_j)) + 3)/4 \\
    A(B_i, B_j) & = \frac{|Area(B1) - Area(B2)|}{area(B1) + Area(B2)}
\end{split}
\end{equation}

After dense prediction and tracking, resulting tracks were all visually examined and corrected in CVAT~\cite{CVAT_ai_Corporation_Computer_Vision_Annotation_2023} when necessary. Specifically, tracks were corrected for identity switches and duplicated or lost tracks. We also removed any false positive tracks (\eg a rock), yielding a total $2139$ animal tracks.

\begin{figure*}
    \centering
    \includegraphics[width=\linewidth]{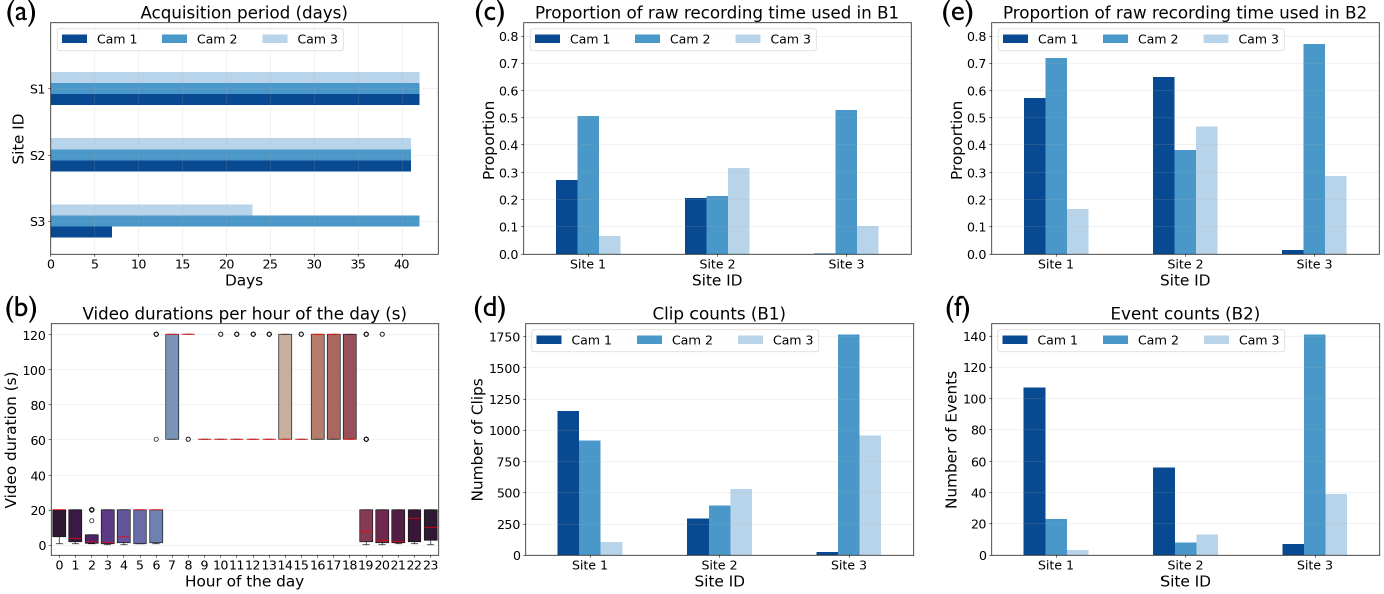}
    \caption{\textbf{Dataset statistics on the acquired data per camera, and on the data used in Sec. {\color{cvprblue} 3.3} and {\color{cvprblue} 3.4}.} (a) summarizes the number of recording days before curation. Note that the batteries for two of the cameras at site S3 ran out earlier. Video durations per hour of the day (b) were computed on the subset of raw videos belonging to either benchmark B1 or B2.}
    \label{fig:acquisition_stats}
\end{figure*}

A video tracklet of dimension $380\times 380$ was created for each individual track by cropping the original video and padding it with the background to preserve the 1:1 aspect ratio.
In crowded scenes, it is common that multiple animals expressing different behaviors are visible on the same tracklet, which may ultimately impact model performance.

The curated tracks include five species: red deer (\textit{Cervus elaphus}), roe deer (\textit{Capreolus capreolus}), fox (\textit{Vulpes vulpes}), wolf (\textit{Canis lupus}) and mountain hare (\textit{Lepus timidus}). Other species were not included, either because too few events were captured or because individuals were too small. 

\subsection{Behavior annotations}

We report the list of behaviors used in the study, along with their definitions and their associated actions, which were automatically gathered from the annotations (\cref{tab:behaviors_table}).
We used a mixed approach to select relevant behaviors. First we sourced behaviors from ethogram studies of related deer species. 
Then, we adjusted the list based on what was interpretable from video-data, and the behavior observed in our data. The \textit{ruminating} behavior was discarded since it was difficult to detect, especially at nighttime, and was hence merged with \textit{standing head up}. The \textit{exploring} behavior was also difficult to differentiate from others, and thus merged with \textit{foraging}. Some social behaviors such as \textit{parenting} or other non-agonistic behaviors between individuals were not included as they are relatively difficult to define in space and time in a consistent manner.

\begin{table*}[]
    \centering
    \begin{NiceTabular}{p{1.5cm}|p{.37\textwidth}|p{.43\textwidth}}
        \toprule
        \textbf{Activity} & \textbf{Associated actions} & \textbf{Definition} \\
        \midrule
         Camera Reaction & standing head up, looking at the camera, running, sniffing, jumping, walking & Any type of behavior that involves reacting to a camera. \\
         \addlinespace[1ex]
         Chasing & running, walking & Whenever a predator chases a prey. \\
         \addlinespace[1ex]
         Courtship & standing head up, running, vocalizing, bathing, scratching antlers, laying, walking & Behaviors related to breeding, uniquely for red deer at this period of the year. It can involve a single stag (\eg vocalizing) or multiple individuals (\eg running after a hind).  \\
         \addlinespace[1ex]
         Escaping & running, vocalizing, walking, jumping & Escaping from a predator, or running away from another individual from the same species. \\
         \addlinespace[1ex]
         Foraging & standing head up, laying, unknown, running, drinking, sniffing, vocalizing, standing head down, bathing, defecating, grazing, walking, urinating, scratching body & Large family of behaviors related to energy acquisition, from environment sensing (\eg sniffing) to actual consumption (\eg grazing). \\ 
         \addlinespace[1ex]
         Grooming & standing head up, shaking fur, bathing, standing head down, scratching antlers, defecating, scratching hoof, laying, walking, urinating, scratching body & Behaviors involving a single individual that cleans its body and fur, either by scratching in multiple ways or while bathing. \\ 
         \addlinespace[1ex]
         Marking & standing head up, defecating, bathing, scratching antlers, standing head down, jumping, scratching hoof, walking, urinating & Behaviors related to a single stag that marks specific features from the environment. \\
         \addlinespace[1ex]
         Playing & standing head up, running, sniffing, standing head down, jumping, scratching hoof, walking & Behaviors involving one or multiple individuals, often young ones, and characterized by running or jumping in the absence of negative stimuli. \\ 
         \addlinespace[1ex]
         Resting & standing head up, bathing, scratching antlers, standing head down, laying & Whenever an animal stays in place for a long time and does not appear to be in vigilance or foraging. \\
         \addlinespace[1ex]
         Unknown & standing head up, unknown, running, sniffing, standing head down, jumping, scratching hoof, walking & Sometimes the behavior cannot be deduced from the current context, for example, because of occlusion or some decisive parts of the body being out-of-frame. \\
         \addlinespace[1ex]
         Vigilance & standing head up, looking at the camera, running, sniffing, standing head down, defecating, grazing, walking & Any behavior where an animal or a group of animals are actively sensing the environment either to detect potential predators or other sources of threat, or in reaction to another individual's vocalization. \\

         \bottomrule
    \end{NiceTabular}
    \caption{\textbf{Definition of the activities present in the dataset and their associated actions.}}
    \label{tab:behaviors_table}
\end{table*}

\subsection{Reference scene segmentation maps}
\begin{figure*}[t] 
    \centering
    \begin{minipage}{0.48\linewidth}
        \centering
        \includegraphics[width=\linewidth]{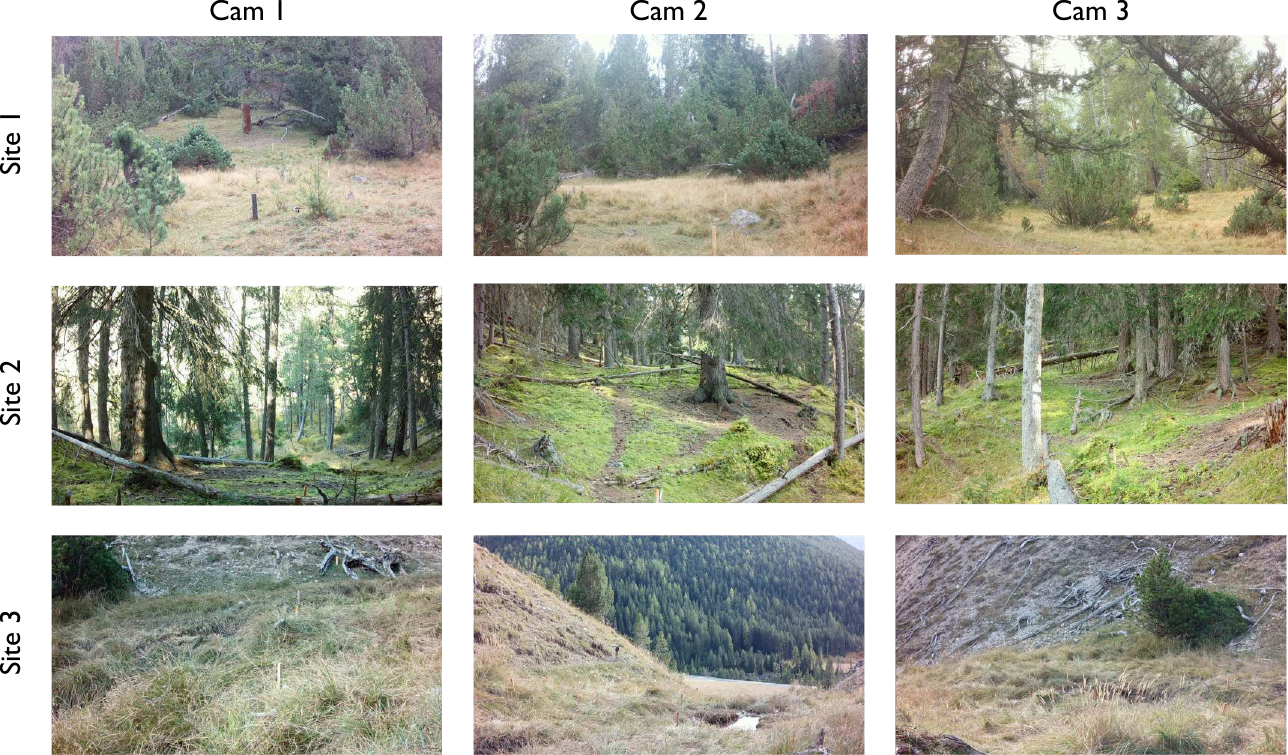}
        \caption{\textbf{Reference picture of the scene for each camera.}}
        \label{fig:ref_scenes}
    \end{minipage}
    \hfill
    \begin{minipage}{0.48\linewidth}
        \centering
        \includegraphics[width=\linewidth]{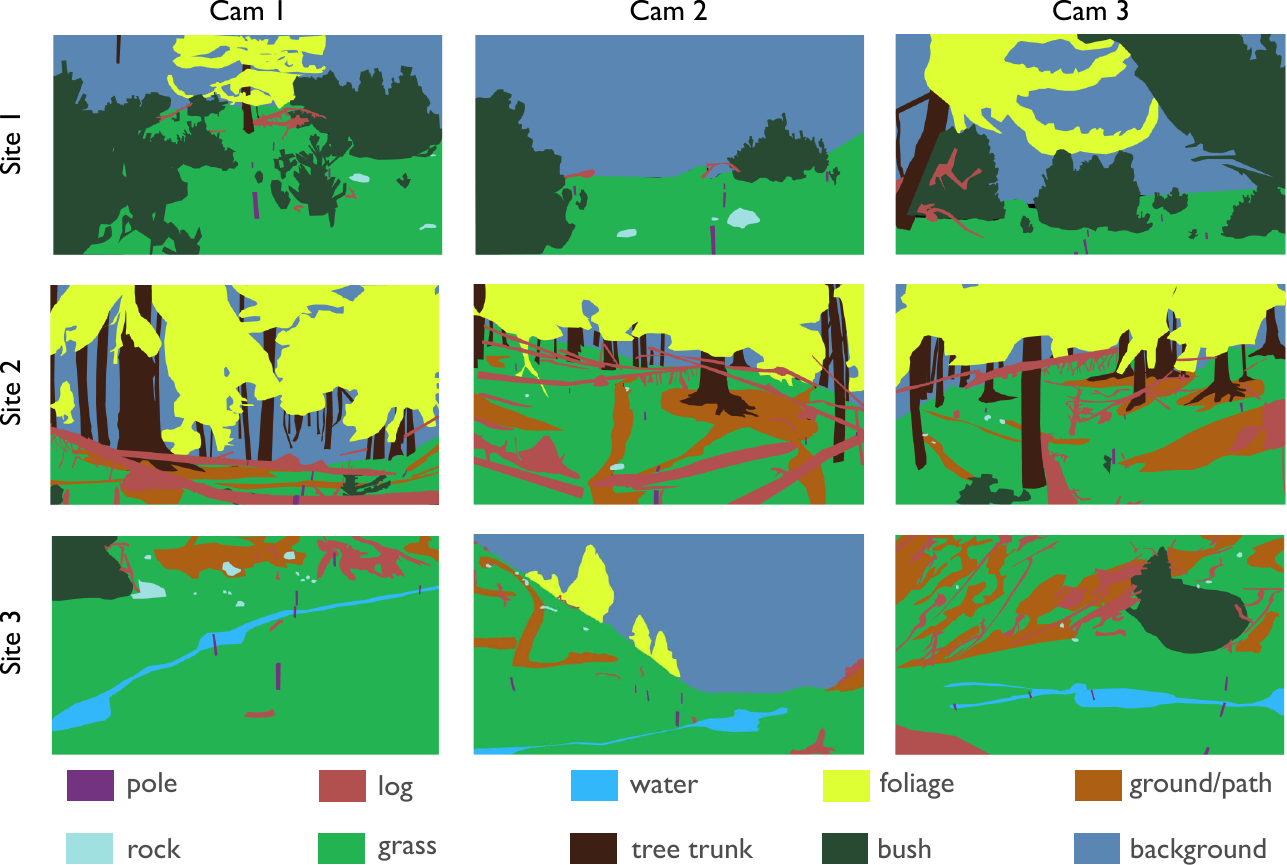}
        \caption{\textbf{Reference scene segmentation maps.}}
        \label{fig:ref_scenes_annot}
    \end{minipage}
\end{figure*}

Before dismounting cameras, a reference picture of the scene was collected for each of them by manually triggering the camera trap (\cref{fig:ref_scenes}). 

The reference scenes (\cref{fig:ref_scenes}) were annotated in CVAT~\cite{CVAT_ai_Corporation_Computer_Vision_Annotation_2023} for 10 classes: \textit{bush}, \textit{pole}, \textit{rock}, \textit{grass}, \textit{soil/path}, \textit{log}, \textit{tree trunk}, \textit{foliage}, \textit{water}, and \textit{background} (\cref{fig:ref_scenes_annot}).

\subsection{Quantification of cameras temporal drift}

\begin{figure*}
    \centering
    \includegraphics[width=0.75\linewidth]{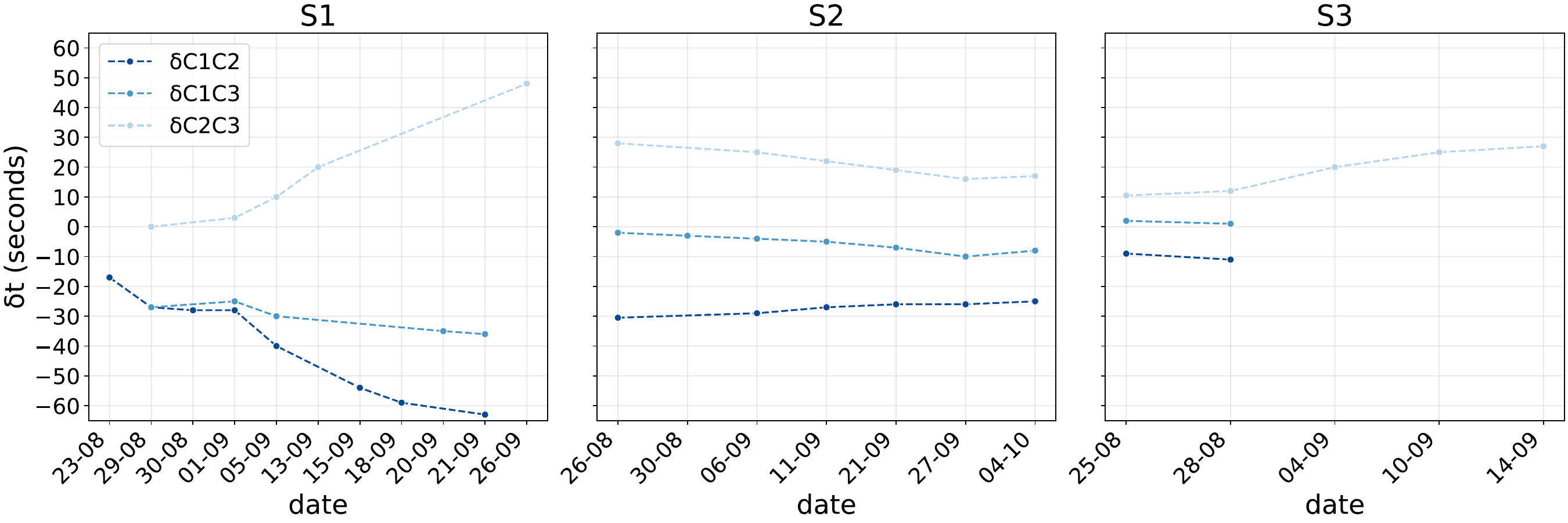}
    \caption{\textbf{Temporal drift between pairs of cameras over time.}}
    \label{fig:camera_drift}
\end{figure*}

We quantified the temporal drift between pairs of cameras for each site, as shown in~\cref{fig:camera_drift}. This was achieved by manually selecting frames that depicted the same animal pose from at least two camera views, and reporting the date and time of the respective frames.
Site 1 shows the biggest drift, while cameras in Site 2 seems less prone to temporal drift. Site 3 contains limited data as Camera 1 battery ran off early, which limits the quantification of the drift.

\clearpage
\section{Benchmark 1: Multimodal Species and Behavior recognition}
\balance

\subsection{Multimodal VideoMAE Implementation details}

We adopted a condensed version of VideoMAE \cite{tong2022videomae} from InternVideo \cite{wang2022internvideo}, for which we used the pre-trained weights on Kinetics 700 dataset \cite{carreira2019short}. We replaced the original classification head with three classification heads to predict species (Spe), activities (ActY) and actions (ActN) simultaneously, while using the loss weights of 1, 2.5 and 2. Meanwhile, we implemented a balanced sampling strategy to deal with the unbalanced number of samples across different classes. For all the models with different modality inputs, we trained them with 150 epochs with the learning rate decreasing from $10^{-5}$ to $10^{-7}$.

\begin{figure}[b]
    \centering
    \includegraphics[width=\linewidth]{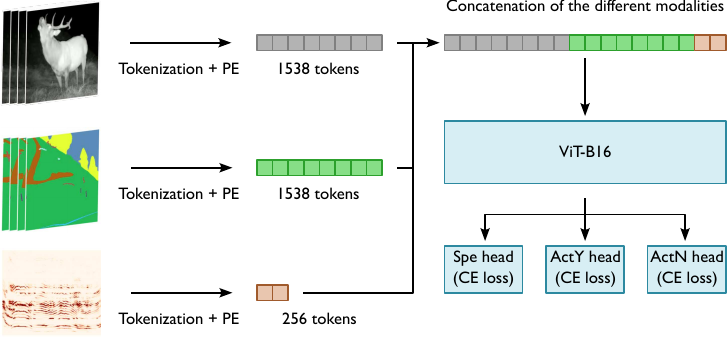}
    \caption{\textbf{Multimodal Video Transformer implementation for B1.} The transformer backbone is similar to both B1 and B2. In B2, the backbone is followed by four classification heads instead of the three depicted here, one for each of the classification tasks.}
    \label{fig:videomae_b1}
\end{figure}

An overview of the model trained for B1 was created (~\cref{fig:videomae_b1}). We made several modifications so that the VideoMAE \cite{tong2022videomae} model can take different modalities as input (video, audio and segmentation masks). First, the video modality is naturally trivial -- we sampled 16 frames similar to the original VideoMAE \cite{tong2022videomae} and then transformed them to $16\times14\times14$ patches. It needs to be noted that we only sampled frames within 5 seconds of randomly selected windows since some behaviors span long times; this captured evidence more compactly. For the audio inputs, we first found the audio clip simultaneous to the video clip and then transformed the original audio signal to a spectrogram, similar to AudioMAE~\cite{huang2022masked}. We adopted a smaller audio sample length (10 in comparison to the original 25) so that the spectrogram can be generated with fewer audio samples. We applied masking across temporal and frequency domains during training for data augmentation. The spectrogram was interpolated to 256 tokens to obtain the same input length across different samples. Finally, for the segmentation inputs, we sampled 16 frames simultaneous to the sampled video frames. Segmentation inputs were represented as one-hot encoded matrices for every frame so that the model did not rely on spurious linear dependencies between the class indices. 

We optimized model parameters by back-propagating the three task-specific cross-entropy (CE) losses. After the quantitative comparison between binary cross-entropy and CE loss for ActN recognition, CE ultimately increased optimization speed, most likely since there are at most two actions and often only one. For both B1 and B2 we used balancing sampling. To account for multiple labels, we computed a sampling weight proportional to the \textit{sum} of their inverse class frequencies.

\subsection{Baseline performance and variability.}

To contextualize the difficulty of B1, we ran additional experiments on the ActY recognition task for videos (\cref{tab:b1_baselines}). Note that the model evaluated on KABR~\cite{kholiavchenko2024kabr} and MammalNet~\cite{chen2023mammalnet} show behavior recognition scores of 0.66 (mAP on X3D-L) and 0.378 (top-1 balanced acc. on mViTv2), respectively, indicating that the difficulty is in the range of related datasets for this single unimodal task.

\begin{table}[!h]
    \centering
    {
    \footnotesize    
    \begin{tabular}{l c c}
    \toprule
        Baseline & mAP & top-1 balanced accuracy \\
    \midrule
        SlowFast-8x8$^\dagger$ & 0.203 & 0.197 \\
        X3D-M$^\dagger$ & 0.251 & 0.256\\
        mViT-v2$^\dagger$ & 0.259 & 0.156 \\
        VideoMAE$^\dagger$ (ours) & 0.410 & 0.274  \\
        VideoMAE (ours) & 0.414 & 0.403\\
    \bottomrule
    \end{tabular}
    }
    \caption{Additional baseline performances on the ActY recognition task from videos. $\dagger$: uniform sampling}
    \label{tab:b1_baselines}
    
\end{table}

\begin{table}[!h]
    \centering
    {\footnotesize
    \begin{tabular}{ l c c c c}
    \toprule
        Mod. & Spe. & ActY. & ActN. & Avg. \\
    \midrule
        V & 0.495 \tiny{$\pm$ 0.020}& 0.436 \tiny{$\pm$ 0.016}& 0.452 \tiny{$\pm$ 0.014}& 0.453 \tiny{$\pm$ 0.002} \\ 
        S & 0.441 \tiny{$\pm$ 0.050}& 0.234 \tiny{$\pm$ 0.027}& 0.172 \tiny{$\pm$ 0.010}& 0.230 \tiny{$\pm$ 0.014} \\ 
        A & 0.223 \tiny{$\pm$ 0.014}& 0.212 \tiny{$\pm$ 0.010}& 0.172 \tiny{$\pm$ 0.002}& 0.192 \tiny{$\pm$ 0.004} \\ 
        V+S & 0.466 \tiny{$\pm$ 0.005}& 0.409 \tiny{$\pm$ 0.009}& 0.384 \tiny{$\pm$ 0.018}& 0.403 \tiny{$\pm$ 0.006} \\ 
        A+S & 0.385 \tiny{$\pm$ 0.012}& 0.312 \tiny{$\pm$ 0.014}& 0.276 \tiny{$\pm$ 0.017}& 0.303 \tiny{$\pm$ 0.012} \\ 
        V+A & 0.473 \tiny{$\pm$ 0.013}& 0.484 \tiny{$\pm$ 0.036}& \textbf{0.466} \tiny{$\pm$ 0.011}& \textbf{0.473} \tiny{$\pm$ 0.017} \\ 
        V+A+S & \textbf{0.531} \tiny{$\pm$ 0.018}& \textbf{0.485} \tiny{$\pm$ 0.014}& 0.437 \tiny{$\pm$ 0.011} & 0.466 \tiny{$\pm$ 0.007} \\ 
    \bottomrule
    \end{tabular}
    }
    \caption{{\bf Variability of the mAPs for the joint task predictions of B1} Mean and standard deviation are computed after training the model three times with different seeds.}
    \label{tab:b1_variability}
\end{table}

\subsection{Models performance per class}

We report model performances (F1-scores and average precisions) per class (\cref{tab:sup_f1_activity}, \cref{tab:sup_map_activity},
\cref{tab:sup_f1_actions} and \ref{tab:sup_map_actions}). The advantage of reporting the mAP (or AP when considering single classes) is that the metric better represents the area under the curve as it computes the precision over multiple thresholds, and it can be equally applied to multi-class and multi-label problems. To compute the F1-score, we used a threshold of 0.5 on the softmax and sigmoid outputs for multi-class and multi-label tasks, respectively.

\begin{table*}[!h]
    \centering
    \small
    \begin{tabular}{l|l|c c c c c c c c c c}
    \toprule
        Activity & Support & \multicolumn{10}{c}{F1-score} \\
    \midrule
        Trained on &       & ActY. & ActY.+ActN. & ActY.+Spe. & All & All & All & All & All & All & All  \\
        Modality   &       & V     & V           & V          & V   & A   & S   & A+S & V+S & V+A & V+A+S\\
    \midrule
        Cam. reaction & 7 & 0.167 & 0.182 & 0.000 & 0.000 & 0.080 & 0.000 & 0.000 & 0.111 & 0.000 & \textbf{0.190} \\
        Chasing & 3 & \textbf{1.000} & \textbf{1.000} & 0.857 & \textbf{1.000} & 0.000 & 0.462 & 0.250 & \textbf{1.000} & 0.857 & 0.750  \\
        Courtship & 56 & 0.565 & 0.532 & 0.429 & 0.442 & 0.589 & 0.143 & 0.512 & 0.330 & 0.574 & \textbf{0.617}\\
        Escaping & 1 & 0.000 & 0.000 & 0.000 & 0.000 & 0.000 & 0.000 & 0.000 & 0.000 & 0.000 & 0.000 \\
        Foraging & 688 & 0.782 & 0.795 & 0.760 & 0.801 & 0.677 & 0.651 & 0.709 & 0.783 & \textbf{0.822} & 0.789 \\
        Grooming & 24 & 0.350 & \textbf{0.359} & 0.264 & 0.293 & 0.014 & 0.108 & 0.150 & 0.230 & 0.356 & 0.310 \\
        Marking & 76 & 0.667 & 0.583 & 0.504 & 0.569 &0.509 & 0.230 & 0.382 & 0.516 & 0.775 & \textbf{0.787} \\
        Playing & 21 & 0.000 & 0.000 & 0.000 & 0.000 & \textbf{0.067} & 0.049 & 0.000 & 0.000 & 0.000 & 0.000 \\
        Resting & 48 & \textbf{0.250} & 0.185 & \textbf{0.250} & 0.189 & 0.000 & 0.039 & 0.000 & 0.207 & 0.154 & 0.185 \\
        Unknown & 92 & 0.426 & 0.398 & 0.394 & 0.378 & 0.030 & 0.275 & 0.229 & 0.441 & \textbf{0.508} & 0.393 \\
        Vigilance & 228 & 0.625 & \textbf{0.664} & 0.619 & 0.637 & 0.025 & 0.183 & 0.338 & 0.589 & 0.640 & 0.621 \\
    \midrule 
        Macro & 1244 & \textbf{0.439} & 0.427 & 0.371 & 0.392 & 0.181 & 0.194 & 0.234 & 0.382 & 0.426 & 0.422
    \end{tabular}
    \caption{\textbf{F1-scores per activity for the behavior recognition benchmark (B1).} V: video clips; A: audio spectrograms; S: segmentation map clips;  ActY.: Activities; ActN.: Actions; Spe.: Species.}
    \label{tab:sup_f1_activity}
\end{table*}

\begin{table*}[!h]
    \centering
    \small
    \begin{tabular}{l|l|c c c c c c c c c c}
    \toprule
        Activity & Support & \multicolumn{10}{c}{AP} \\
    \midrule
        Trained on &       & ActY. & ActY.+ActN. & ActY.+Spe. & All & All & All & All & All & All & All  \\
        Modality   &       & V     & V           & V          & V   & A   & S   & A+S & V+S & V+A & V+A+S\\
    \midrule
        Cam. reaction & 7 & 0.089 & 0.114 & 0.169 & 0.119 & 0.018 & 0.042 & 0.073 & 0.104 & 0.114 & \textbf{0.194}  \\
        Chasing & 3 & \textbf{1.000} & \textbf{1.000} & \textbf{1.000} & \textbf{1.000} & 0.017 & 0.362 & 0.423 & \textbf{1.000} & \textbf{1.000} & 0.917 \\
        Courtship & 56 & 0.540 & 0.552 & 0.425 & 0.419 & 0.638 & 0.113 & 0.569 & 0.369 & 0.633 & 0.\textbf{651} \\
        Escaping & 1 & 0.059 & 0.034 & \textbf{0.333} & 0.023 & 0.006 & 0.004 & 0.015 & 0.077 & 0.017 & 0.038 \\
        Foraging & 688 & 0.850 & 0.870 & 0.857 & 0.867 & 0.613 & 0.703 & 0.735 & 0.840 & \textbf{0.873} & 0.870 \\
        Grooming & 24 & 0.280 & 0.291 & 0.216 & \textbf{0.308} & 0.020 & 0.101 & 0.116 & 0.152 & 0.307 & 0.222\\
        Marking & 76 & 0.739 & 0.619 & 0.572 & 0.654 & 0.534 & 0.155 & 0.321 & 0.556 & \textbf{0.794} & 0.788 \\
        Playing & 21 & 0.017 & 0.022 & 0.026 & 0.024 & 0.042 & \textbf{0.071} & 0.050 & 0.055 & 0.036 & 0.030\\
        Resting & 48 & 0.275 & \textbf{0.289} & 0.286 & 0.267 & 0.070 & 0.051 & 0.068 & 0.205 & 0.280 & 0.218 \\
        Unknown & 92 & 0.342 & 0.367 & 0.344 & 0.357 & 0.104 & 0.227 & 0.208 & 0.421 & \textbf{0.456} & 0.395 \\
        Vigilance & 228 & 0.651 & 0.706 & 0.646 & 0.672 & 0.218 & 0.241 & 0.310 & 0.608 & \textbf{0.713} & 0.651 \\
    \midrule 
        Macro & 1244 & 0.440 & 0.442 & 0.443 & 0.428 & 0.207 & 0.188 & 0.262 & 0.399 & \textbf{0.475} & 0.452
    \end{tabular}
    \caption{\textbf{Average precisions (AP) per activity for the behavior recognition benchmark (B1).} V: video clips; A: audio spectrograms; S: segmentation map clips;  ActY.: Activities; ActN.: Actions; Spe.: Species.}
    \label{tab:sup_map_activity}
\end{table*}

\begin{table*}[!htb]
    \centering
    \small
    \begin{tabular}{l|l|c c c c c c c c c c}
    \toprule
        Action & Support & \multicolumn{10}{c}{F1-score} \\
    \midrule
        Trained on &       & ActN. & ActY.+ActN. & ActN.+Spe. & All & All & All & All & All & All & All  \\
        Modality   &       & V     & V           & V          & V   & A   & S   & A+S & V+S & V+A & V+A+S\\
    \midrule
        Bathing         &2   & \textbf{0.400} & 0.286 & \textbf{0.400} & \textbf{0.400} & 0.013 & 0.028 & 0.071 & 0.133 & 0.286 & \textbf{0.400} \\
        Defecating      &6   & 0.000 & 0.000 & 0.000 & 0.000 & 0.026 & 0.022 & \textbf{0.040} & 0.000 & 0.000 & 0.013 \\
        Drinking        &6   & \textbf{0.500} & 0.444 & 0.400 & 0.444 & 0.033 & 0.062 & 0.156 & 0.267 & 0.345 & 0.316 \\
        Grazing         &184 & \textbf{0.684} & 0.613 & 0.650 & 0.616 & 0.425 & 0.510 & 0.508 & 0.564 & 0.592 & 0.564 \\
        Jumping         &7   & 0.000 & 0.000 & \textbf{0.222} & 0.000 & 0.044 & 0.108 & 0.000 & 0.000 & 0.143 & 0.000 \\
        Laying          &53  & 0.312 & \textbf{0.435} & 0.394 & 0.317 & 0.102 & 0.062 & 0.051 & 0.314 & 0.344 & 0.303 \\
        Look. at cam.   &2   & 0.000 & 0.000 & 0.000 & \textbf{0.333} & 0.000 & 0.041 & 0.118 & 0.074 & 0.000 & 0.000 \\
        Running         &36  & 0.466 & 0.416 & 0.376 & \textbf{0.471 }& 0.162 & 0.305 & 0.325 & 0.313 & 0.455 & 0.330 \\
        Scratch. antlers&55  & 0.638 & 0.645 & 0.626 & 0.680 & 0.280 & 0.188 & 0.258 & 0.508 & \textbf{0.745} & 0.686 \\
        Scratch. body   &10  & \textbf{0.250} & 0.187 & 0.211 & 0.000 & 0.000 & 0.015 & 0.030 & 0.083 & 0.091 & 0.139 \\
        Scratch. hoof   &24  & 0.294 & 0.321 & 0.373 & 0.280 & 0.236 & 0.127 & 0.286 & 0.200 & 0.429 & \textbf{0.430} \\
        Shaking fur     &11  & 0.545 & \textbf{0.571} & 0.400 & 0.538 & 0.020 & 0.101 & 0.161 & 0.359 & 0.273 & 0.350 \\
        Sniffing        &38  & \textbf{0.479} & 0.143 & 0.232 & 0.193 & 0.064 & 0.081 & 0.116 & 0.120 & 0.218 & 0.118 \\
        Stand. head down&180 & 0.464 & 0.375 & \textbf{0.467} & 0.400 & 0.279 & 0.300 & 0.298 & 0.385 & 0.400 & 0.381 \\
        Stand. head up  &265 & 0.689 & \textbf{0.712} & 0.648 & 0.677 & 0.359 & 0.390 & 0.397 & 0.585 & 0.702 & 0.629 \\
        Unknown         &75  & \textbf{0.578} & 0.551 & 0.507 & 0.497 & 0.147 & 0.283 & 0.217 & 0.357 & 0.502 & 0.401 \\
        Urinating       &1   & 0.000 & 0.000 & 0.000 & 0.000 & 0.000 & 0.000 & 0.000 & 0.000 & 0.000 & 0.000 \\
        Vocalizing      &37  & 0.323 & 0.500 & 0.328 & 0.505 & \textbf{0.604} & 0.188 & 0.481 & 0.306 & 0.598 & 0.511 \\  
        Walking         &300 & \textbf{0.786} & 0.746 & 0.780 & 0.714 & 0.400 & 0.458 & 0.491 & 0.548 & 0.730 & 0.658 \\
    \midrule 
        Macro & 1292*        & \textbf{0.390} & 0.366 & 0.369 & 0.372 & 0.168 & 0.172 & 0.211 & 0.269 & 0.361 & 0.328 \\
    \end{tabular}
    \caption{\textbf{F1-scores per action for the behavior recognition benchmark (B1).} *Note that since there can be up to two actions per sample, this increases the total number of samples since each label is considered independently. V: video clips; A: audio spectrograms; S: segmentation map clips;  ActY.: Activities; ActN.: Actions; Spe.: Species.}
    \label{tab:sup_f1_actions}
\end{table*}

\begin{table*}[!htb]
    \centering
    \small
    \begin{tabular}{l|l|c c c c c c c c c c}
    \toprule
        Action & Support & \multicolumn{10}{c}{AP} \\
    \midrule
        Trained on &       & ActY. & ActY.+ActN. & ActY.+Spe. & All & All & All & All & All & All & All  \\
        Modality   &       & V     & V           & V          & V   & A   & S   & A+S & V+S & V+A & V+A+S\\
        \midrule
        Bathing         &2   & 0.507 & 0.509 & 0.528 & \textbf{0.550} & 0.011 & 0.254 & 0.508 & 0.503 & 0.520 & 0.507 \\
        Defecating      &6   & 0.008 & 0.012 & 0.008 & 0.006 & 0.014 & 0.007 & \textbf{0.061} & 0.005 & 0.005 & 0.007 \\
        Drinking        &6   & 0.633 & 0.555 & 0.714 & 0.513 & 0.051 & 0.029 & 0.166 & \textbf{0.800} & 0.621 & 0.502 \\
        Grazing         &184 & \textbf{0.857} & 0.746 & 0.848 & 0.847 & 0.388 & 0.493 & 0.544 & 0.792 & 0.834 & 0.812 \\
        Jumping         &7   & 0.023 & 0.023 & \textbf{0.207} & 0.024 & 0.032 & 0.042 & 0.043 & 0.014 & 0.105 & 0.024 \\
        Laying          &53  & 0.315 & 0.368 & 0.381 & 0.369 & 0.054 & 0.085 & 0.093 & 0.244 & \textbf{0.382} & 0.321 \\
        Look. at cam.   &2   & 0.008 & 0.013 & 0.012 & \textbf{0.238} & 0.002 & 0.126 & 0.035 & 0.026 & 0.047 & 0.030 \\
        Running         &36  & 0.669 & \textbf{0.684} & 0.586 & 0.634 & 0.172 & 0.315 & 0.457 & 0.489 & 0.646 & 0.521 \\
        Scratch. antlers&55  & 0.674 & 0.672 & 0.654 & 0.716 & 0.184 & 0.160 & 0.192 & 0.558 & 0.742 & \textbf{0.760} \\
        Scratch. body   &10  & \textbf{0.164} & 0.091 & 0.152 & 0.067 & 0.009 & 0.014 & 0.017 & 0.044 & 0.054 & 0.097 \\
        Scratch. hoof   &24  & 0.294 & 0.212 & 0.198 & 0.292 & 0.289 & 0.069 & 0.299 & 0.166 & 0.470 & \textbf{0.519} \\
        Shaking fur     &11  & \textbf{0.559} & 0.400 & 0.323 & 0.516 & 0.020 & 0.134 & 0.126 & 0.248 & 0.345 & 0.272 \\
        Sniffing        &38  & \textbf{0.517} & 0.320 & 0.399 & 0.456 & 0.037 & 0.101 & 0.122 & 0.246 & 0.407 & 0.167 \\
        Stand. head down&180 & 0.575 & 0.477 & \textbf{0.578} & 0.535 & 0.234 & 0.197 & 0.181 & 0.313 & 0.521 & 0.346 \\
        Stand. head up  &265 & 0.778 & \textbf{0.853} & 0.806 & 0.851 & 0.035 & 0.362 & 0.462 & 0.772 & 0.830 & 0.806 \\
        Unknown         &75  & 0.610 & 0.607 & \textbf{0.619} & 0.572 & 0.093 & 0.280 & 0.278 & 0.507 & 0.576 & 0.519 \\
        Urinating       &1   & \textbf{0.007} & 0.002 & \textbf{0.007} & 0.003 & 0.002 & 0.005 & 0.002 & 0.004 & 0.001 & 0.003 \\
        Vocalizing      &37  & 0.415 & 0.688 & 0.500 & 0.606 & 0.835 & 0.100 & 0.724 & 0.561 & 0.787 & \textbf{0.836} \\  
        Walking         &300 & 0.890 & 0.881 & 0.876 & \textbf{0.901} & 0.284 & 0.477 & 0.569 & 0.841 & 0.895 & 0.878 \\
    \midrule 
        Macro & 1292*   & 0.447 & 0.427 & 0.442 & 0.458 & 0.161 & 0.171 & 0.257 & 0.375 & \textbf{0.463} & 0.417 \\
    \end{tabular}
    \caption{\textbf{Average precisions (AP) per action for the behavior recognition benchmark (B1).} *Note that since there can be up to two actions per sample, this increases the total number of samples since each label is considered independently. V: video clips; A: audio spectrograms; S: segmentation map clips;  ActY.: Activities; ActN.: Actions; Spe.: Species.}
    \label{tab:sup_map_actions}
\end{table*}

\clearpage
\section{Benchmark 2: Multi-view Long-term Event Understanding}

Here we detail our simple baseline method for B2. In particular, we illustrate how we performed token merging, how we trained the model and additional results.  

\subsection{Selecting false positive events}
The raw video dataset contains 43 h of raw data, where the majority comes from false positive samples in Camera 1 of site 3 (\cref{fig:acquisition_stats}a-b). While having these false positive events is important for B2 as they represent true data and are common in camera trap surveys, a disproportionate number of them leads to unnecessarily high computational costs.
To construct the dataset for B2, we therefore discarded any event that was longer than 15 minutes (cumulative recording time among all points of view) which eliminated 10 false positive events and three true positive ones, and effectively reducing the dataset size to 14 hours with 3 hours of false positive events.

\subsection{Offline Token Merging strategy}
\begin{figure*}
    \centering
    \includegraphics[width=\linewidth]{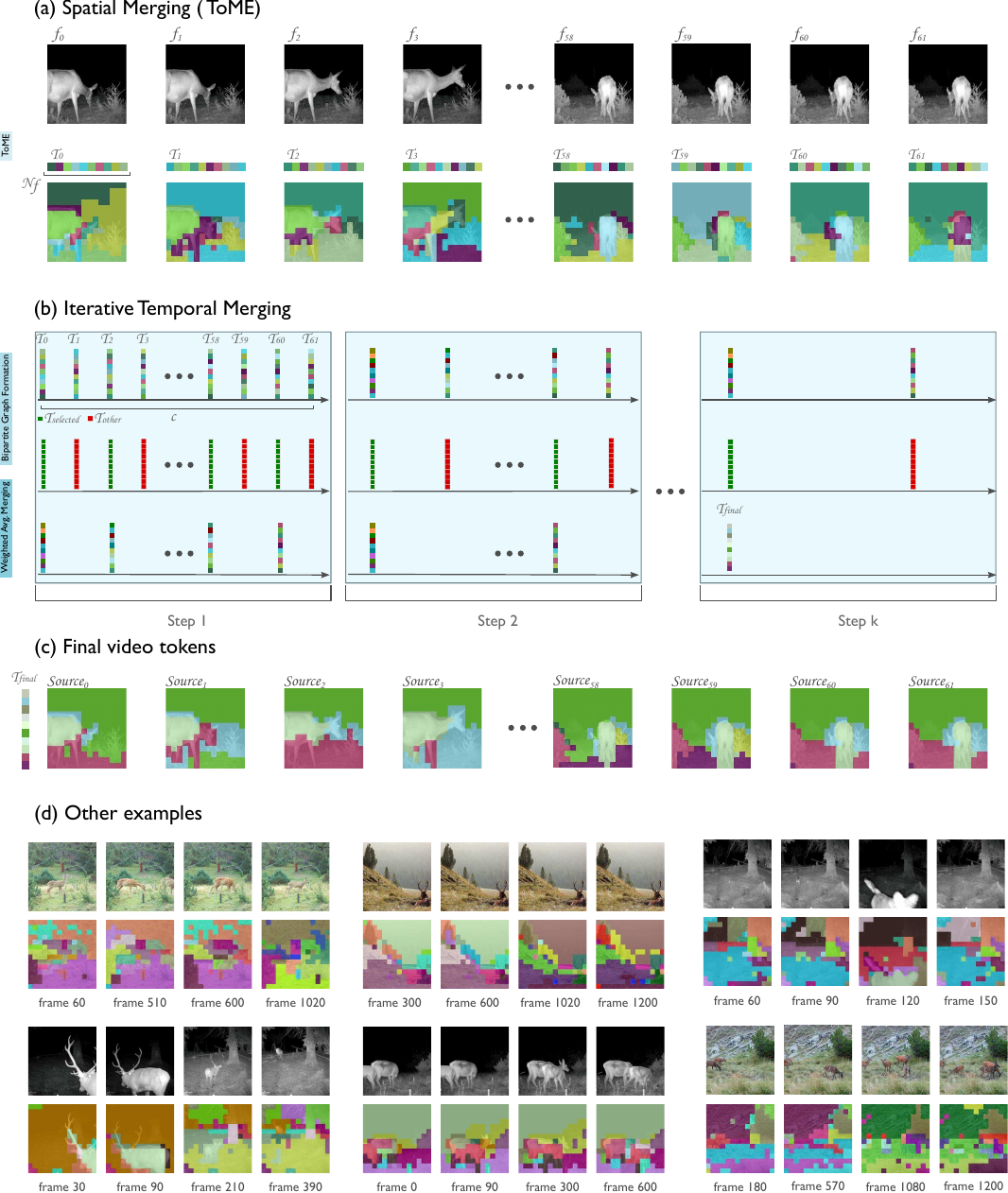}
    \caption{\textbf{Offline token merging strategy.} (a) We apply ToME~\cite{bolya2022token} first spatially and then (b-c) temporally. (d) Multiple examples show initial frames and the source patch and frames for each final video token, corresponding to a unique color. For (a-c), refer to~\cref{alg:token_merging} for variable names. For (a-b) we use a ToME reduction factor of 16, for (d) we use a ToME reduction factor of 14.}
    \label{fig:token_merging}
\end{figure*}

We describe our offline token merging strategy over time in ~\cref{alg:token_merging}, and illustrate the process (~\cref{fig:token_merging}).
After spatial merging with ToME~\cite{bolya2022token}, we select the tokens of every second frame and merge them with any other tokens from all the other frames, following the same soft-bipartite graph matching algorithm used in the original method~\cite{bolya2022token}.
The process is repeated iteratively the final number of video tokens is equal or inferior to the original number of tokens in a single frame. 
Note that the final number of video tokens increases with the video duration since we perform the algorithm in chunks. The embedding dimension is 768, and the chunk size is 615 frames.

\begin{algorithm}
\small  
\caption{Offline Token Merging}
\label{alg:token_merging}
\begin{algorithmic}[1]
\Require Video frames $\mathcal{F}$,

Pretrained Vision-MAE with token merging~\cite{bolya2022token} ToME,

ToME reduction factor $r$,

Chunk size $c$
\Ensure Condensed video tokens $\mathcal{T}_{\text{final}}$

\For{each chunk $\mathcal{C}_i \subset \mathcal{F}$ of size $c$} \Comment{Process in chunks}
    \State $\mathcal{T} \gets \text{ToME}(f_j, r), \forall f_j \in \mathcal{C}_i$ \Comment{Spatial Merging}
    \State $N_f \gets |\mathcal{T}_{j}| \text{ for any } j$
    \Comment{Tokens in each frame}
    \State $N_v \gets N_f \times |\mathcal{C}_i|$ 
    \Comment{Tokens in chunk}
    \While{$N_v > N_f$} \Comment{Temporal Merging}
        \State $\mathcal{T}_{\text{selected}} \gets \{ \mathcal{T}_j \mid j \text{ is even} \}$
        \State $\mathcal{T}_{\text{other}} \gets \{ \mathcal{T}_j \mid j \text{ is odd} \}$ 
        \State $\mathcal{T} \gets \text{Merge}(\mathcal{T}_{\text{selected}}, \mathcal{T}_{\text{other}})$ 
        \State $\mathcal{C}_i \gets \{ f_j \mid j \text{ is even}, \forall f \in \mathcal{C}_i \}$
        \State $N_v \gets N_f \times |\mathcal{C}_i|$
    \EndWhile
\EndFor

\State $\mathcal{T}_{\text{final}} \gets \bigcup_i \mathcal{T}_{\mathcal{C}_i}$ \Comment{Concatenate tokens across chunks}

\end{algorithmic}
\normalsize  
\end{algorithm}

\subsection{Transformer encoder implementation details}

We used the same code base as for B1 for the long-term event understanding task. Instead of giving video frames to a video tokenizer as input to a transformer encoder, we concatenated all video tokens corresponding to a given event, while adding spatial ($Cam_{ID}$: camera id) and positional encodings ($\Delta T_{event}$: elapsed time w.r.t event start), and input them to the transformer encoder. We also added the source frame and patches from the offline token merging process to each individual video token as positional embedding ($Source$). We used the same encoder as a ViT-base model, without using pretraining weights (i.e. trained from scratch).

Models were trained for 300 epochs with a learning rate decreasing from $10^{-5}$ to $10^{-7}$ using the Adam-weighted optimizer. We applied the same sampling balancing strategy as in B1. We trained the activity recognition task with binary cross-entropy loss, and the other three tasks with categorical cross-entropy loss. We did not apply loss weighting to any of the four classification heads.

\subsection{Camera-views ablation} 
We ablated camera-views: C$_1$, C$_2$ (Table~\ref{tab:b2_ablations}). Models are tested on the same multi-view subset of events E$_{C1}\cup$E$_{C2}$, which are seen by either one or both views. Experiments demonstrate the advantage of using multiple views for complex tasks such as ActY recognition and number of individuals recognition.

\begin{table}[ht!]
    \small{
    \centering
    \begin{tabular}{l c c c}
    \toprule
        Train events & ActY mAP & Ind. mAP & Avg. mAP\\
    \midrule
       E$_{C1}$ & 0.379 & 0.474 & 0.407\\
       E$_{C2}$ & 0.464 & 0.445 & 0.446 \\
       E$_{C1}\cup$E$_{C2}$ $\diagdown$ E$_{C1}\cap$E$_{C2}$ & 0.480 & 0.445 & 0.456 \\
       E$_{C1}\cup$E$_{C2}$ & 0.522 & 0.510 & 0.501 \\
    \bottomrule
    \end{tabular}
    \caption{\textbf{Camera-view ablations for B2.} Models are trained with all positional embeddings and $r=14$ on the joint recognition task. ActY: Activity; Ind. Number of individuals; Avg. Overall per-class }
    \label{tab:b2_ablations}
    }
\end{table}

\begin{table*}[h]
    \centering
    \begin{tabular}{c c c c c c c}
    \toprule
        $r$ & Cont. Len. & Spe. & ActY. & Met. Cond. & Indiv. & Avg. \\
    \midrule
        14 & 4096 & 0.415 \scriptsize{$\pm$ 0.084} & 0.479 \scriptsize{$\pm$ 0.032} & 0.618 \scriptsize{$\pm$ 0.034} & 0.499 \scriptsize{$\pm$ 0.019} & 0.489 \scriptsize{$\pm$ 0.033} \\ 
        11 & 8192 & 0.446 \scriptsize{$\pm$ 0.059} & 0.481 \scriptsize{$\pm$ 0.032} & 0.594 \scriptsize{$\pm$ 0.035} & 0.543 \scriptsize{$\pm$ 0.033} & 0.500 \scriptsize{$\pm$ 0.004} \\ 
    \bottomrule
    \end{tabular}

    \caption{{\bf Variability of the mAPs for the joint task predictions of B2} Mean and standard deviation are computed after training the model three times with different seeds. "$r$": ToME~\cite{bolya2022token} reduction factor. A larger reduction factor leads to more patches being merged at the frame level and fewer video tokens; "Cont. Len.": context length: number of tokens per sample; ActY.: Activities; Spe.: Species.; Met. Cond.: Meteorological Conditions; Indiv.: Number of individuals categories.; Avg.: overall per-class average.}
    \label{tab:b2_variability}
\end{table*}

\subsection{Models performance per class}

We report F1-scores and average precisions per class computed similarly as for B1 (\cref{tab:sup_b2_F1} and \ref{tab:sup_b2_AP}). We show the results when using a ToME~\cite{bolya2022token} reduction factor of $r=14$ and $r=11$, and all types of positional encodings ($Cam_{ID}$, $\Delta T_{event}$, $Source$).

\begin{table}[]
    \centering
    \small
    \begin{tabular}{l|c|c c}
    Class & Support & $r=14$ & $r=11$ \\
    \toprule
    \multicolumn{4}{c}{Activities AP} \\
    \midrule
        Cam. reaction & 5 & \textbf{0.300} & 0.080\\
        Chasing & 2 & 0.022 & \textbf{0.175} \\
        Courtship & 5 &  0.385 & \textbf{0.396} \\
        Escaping & 2 & 0.021 & \textbf{0.038} \\
        Foraging & 49 & 0.863 & \textbf{0.890} \\
        Grooming & 5 & \textbf{0.648} & 0.499 \\
        Marking & 5 & \textbf{0.557} & 0.465 \\
        None & 28 & 0\textbf{.959} & 0.924 \\
        Playing & 1 & \textbf{0.042} & 0.020 \\
        Resting & 3 &  \textbf{0.459} & 0.411 \\
        Unknown & 30 & \textbf{0.786} & 0.755 \\
        Vigilance & 35 & \textbf{0.759} & 0.751 \\
    \midrule
        Macro & 170* & \textbf{0.483} & 0.450 \\
        \addlinespace[2ex]
    \midrule
    \multicolumn{4}{c}{Species AP} \\
    \midrule
        Fox & 1 & 0.030 & \textbf{0.500} \\
        Hare & 1 & \textbf{0.020} & 0.019 \\
        None & 28 & 0.919 & \textbf{0.978} \\
        Red deer & 53 & 0.938 & \textbf{0.973} \\
        Roe deer & 3 &  0.118 & \textbf{0.148} \\
        Wolf & 1 & \textbf{0.033} & 0.018 \\
    \midrule
        Macro & 87* & 0.343 & \textbf{0.439} \\
        \addlinespace[2ex]
    \midrule
    \multicolumn{4}{c}{Meteorological Conditions AP} \\
    \midrule
       Clear & 30 & \textbf{0.803} & 0.798 \\
       Overcast  & 15 & 0.466 & \textbf{0.533} \\
       Rainy & 9 & \textbf{0.416} & 0.348 \\
       Sunny & 32 & \textbf{0.927} & 0.858 \\
    \midrule
       Macro & 86 & \textbf{0.653}  & 0.634 \\
       \addlinespace[2ex]
    \midrule
    \multicolumn{4}{c}{Counting Individuals AP} \\
    \midrule
        0 & 28 & 0.917 & \textbf{0.985} \\
        1 & 42 & 0.684 & \textbf{0.798} \\
        2 & 10 & 0.170 & \textbf{0.348} \\
        3+ & 6 & 0.014 & \textbf{0.239} \\
    \midrule
        Macro & 86 & 0.478 & \textbf{0.593} \\
    \bottomrule
    
    \end{tabular}
    \caption{\textbf{Average precisions (AP) per class for the long-term event understanding benchmark (B2).} *Note that since there can be multiple species and activities per sample, this increases the total support since each label is considered independently.}
    \label{tab:sup_b2_AP}
\end{table}

\begin{table}[]
    \centering
    \small
    \begin{tabular}{l|c|c c}
    Class & $r=14$ & $r=11$ \\
    \toprule
    \multicolumn{4}{c}{Activities F1-scores} \\
    \midrule
        Cam. reaction & 5 & 0.222 & 0.000 \\
        Chasing & 2 & 0.000 & 0.000 \\
        Courtship & 5 & \textbf{0.333} & \textbf{0.333} \\
        Escaping & 2 & 0.000 & 0.000 \\
        Foraging & 49 & \textbf{0.889} & 0.871 \\
        Grooming & 5 & \textbf{0.400} & 0.250 \\
        Marking & 5 & 0.286 & \textbf{0.333} \\
        None & 28 & 0.926 & \textbf{0.964} \\
        Playing & 1 & 0.000 & 0.000 \\
        Resting & 3 & \textbf{0.500} & 0.000 \\
        Unknown & 30 & \textbf{0.812} & 0.704 \\
        Vigilance & 35 & 0.658 & \textbf{0.667} \\
    \midrule
        Macro & 170* & \textbf{0.419} & 0.344 \\
        \addlinespace[2ex]
    \midrule
    \multicolumn{4}{c}{Species F1-scores} \\
    \midrule
        Fox & 1 & 0.000 & 0.000\\
        Hare & 1 & 0.000 & 0.000 \\
        None & 28 & \textbf{0.926} & \textbf{0.926} \\
        Red deer & 53 & \textbf{0.909} & 0.907 \\
        Roe deer & 3 & \textbf{0.222} & 0.182 \\
        Wolf & 1 & 0.000 & 0.000 \\
    \midrule
        Macro & 87* & \textbf{0.343} & 0.336\\
        \addlinespace[2ex]
    \midrule
    \multicolumn{4}{c}{Meteorological Conditions F1-scores} \\
    \midrule
       Clear  & 30 & \textbf{0.778} & \textbf{0.778} \\
       Overcast  & 15 & \textbf{0.545} & 0.300 \\
       Rainy & 9 & \textbf{0.333} & \textbf{0.333} \\
       Sunny & 32 & 0.939 & \textbf{0.941} \\
    \midrule
       Macro & 86 & \textbf{0.649} & 0.588 \\
       \addlinespace[2ex]
    \midrule
    \multicolumn{4}{c}{Counting Individuals F1-scores} \\
    \midrule
        0 & 28 & 0.926 & \textbf{0.964} \\
        1 & 42 & 0.690 & \textbf{0.833} \\
        2 & 10 & \textbf{0.174} & 0.000 \\
        3+ & 6 & 0.000 & 0.000 \\
    \midrule
        Macro & 86 & 0.448 & \textbf{0.449} \\
    \bottomrule
    
    \end{tabular}
    \caption{\textbf{F1-scores per class for the long-term event understanding benchmark (B2).} *Note that since there can be multiple species and activities per sample, this increases the total support since each label is considered independently.}
    \label{tab:sup_b2_F1}
\end{table}

\end{document}